\documentclass[letterpaper]{article} 
\usepackage{aaai23}  
\usepackage{times}  
\usepackage{helvet}  
\usepackage{courier}  
\usepackage[hyphens]{url}  
\usepackage{graphicx} 
\usepackage{natbib}  
\usepackage{caption} 
\usepackage{algorithm}
\usepackage{algorithmic}

\usepackage{newfloat}
\usepackage{listings}
\DeclareCaptionStyle{ruled}{labelfont=normalfont,labelsep=colon,strut=off} 
\floatstyle{ruled}
\newfloat{listing}{tb}{lst}{}
\floatname{listing}{Listing}
\usepackage{tikz}
\usepackage{comment}
\usepackage{amsmath,amssymb}
\usepackage{color}
\usepackage{xcolor}
\usepackage{xspace}

\DeclareMathOperator*{\argmin}{arg\,min}
\usepackage{multirow}
\usepackage{adjustbox}
\usepackage{siunitx}
\usepackage{booktabs}
\usepackage{subfig}
\usepackage{placeins}
\usepackage{pifont}
\usepackage{wrapfig,lipsum}

\newcommand{\cmark}{\ding{51}}
\newcommand{\xmark}{\ding{53}}

\newcommand{\modelname}{\text{Penalized Diversity}\xspace}
\newcommand{\model}{\textsc{PD}\xspace}
\newcommand{\diversitycontrib}{\textsc{DBA}\xspace}
\newcommand{\penaltycontrib}{\textsc{WHP}\xspace}
\newcommand{\penaltycontriblower}{\text{whp}\xspace}
\newcommand{\sharedb}{\text{ShB}\xspace}
\newcommand{\separateb}{\text{SeB}\xspace}

\nocopyright 

\title{Source-free Domain Adaptation Requires Penalized Diversity}
\author {
    Laya Rafiee Sevyeri,\textsuperscript{\rm 1 \rm 2} \footnote{This research was conducted during author’s internship at Imagia.}
    Ivaxi Sheth,\textsuperscript{\rm 2 \rm 3 \rm 4}
    Farhood Farahnak,\textsuperscript{\rm 1}
    Alexandre See,\textsuperscript{\rm 2}
    Samira Ebrahimi Kahou,\textsuperscript{\rm 3 \rm 4 \rm 5}
    Thomas Fevens,\textsuperscript{\rm 1}
    Mohammad Havaei \textsuperscript{\rm 2}
}
\affiliations {
    \textsuperscript{\rm 1} Concordia University, Montr\'eal, Canada \\
    \textsuperscript{\rm 2} Imagia Canexia Health, Montr\'eal, Canada \\
    \textsuperscript{\rm 3} École de Technologie Supérieure, Montr\'eal, Canada \\
    \textsuperscript{\rm 4} Mila, Montr\'eal, Canada \\
    \textsuperscript{\rm 5} CIFAR AI Chair \\
    laya.rafiee@gmail.com, 
    ivaxisheth17@gmail.com, 
    farhood.farahnak@gmail.com, 
    alexandre.see@imagia.com, 
    samira.ebrahimi.kahou@gmail.com,
    thomas.fevens@concordia.ca, 
    mohammad@imagia.com
}
\usepackage{bibentry}

\begin{document}

\maketitle

\begin{abstract}
While neural networks are capable of achieving human-like performance in many tasks such as image classification, the impressive performance of each model is limited to its own dataset. Source-free domain adaptation (SFDA) was introduced to address knowledge transfer between different domains in the absence of source data, thus, increasing data privacy. Diversity in representation space can be vital to a model's adaptability in varied and difficult domains. In unsupervised SFDA, the diversity is limited to learning a single hypothesis on the source or learning multiple hypotheses with a shared feature extractor. Motivated by the improved predictive performance of ensembles, we propose a novel unsupervised SFDA algorithm that promotes representational diversity through the use of separate feature extractors with Distinct Backbone Architectures (DBA). Although diversity in feature space is increased, the unconstrained mutual information (MI) maximization may potentially introduce amplification of weak hypotheses. Thus we introduce the Weak Hypothesis Penalization (\penaltycontrib) regularizer as a mitigation strategy. Our work proposes Penalized Diversity (\model) where the synergy of \diversitycontrib and \penaltycontrib is applied to unsupervised source-free domain adaptation for covariate shift. In addition, \model is augmented with a weighted MI maximization objective for label distribution shift. Empirical results on natural, synthetic, and medical domains demonstrate the effectiveness of \model under different distributional shifts.
\end{abstract}

\section{Introduction}
In recent years, the field of machine learning (ML) has witnessed immense progress in computer vision~\cite{he2016deep}, natural language processing~\cite{vaswani2017transformer}, and speech recognition~\cite{bahdanau2016end} due to the advances of deep neural networks (DNNs). Despite the increasing popularity of DNNs, they often perform poorly on unseen distributions~\cite{geirhos2020shortcut}, leading to overconfident and miscalibrated models. Combining the predictions of several models seems to be a feasible way to improve the generalizability of these models~\cite{turner1999decimated}. On account of its simplicity and effectiveness, ensemble learning became popular in many machine learning applications. Due to the i.i.d. assumption that training and test sets are drawn from the same distribution, calibration~\cite{dawid1982well} is introduced to the traditional machine learning paradigm to elucidate the model uncertainty. Additionally, predictive uncertainty is crucial under dataset shift--when confronted with a sample from a shifted distribution, an ideal model should reflect increased uncertainty in its prediction.
\begin{figure}
\centering
\includegraphics[scale=0.2]{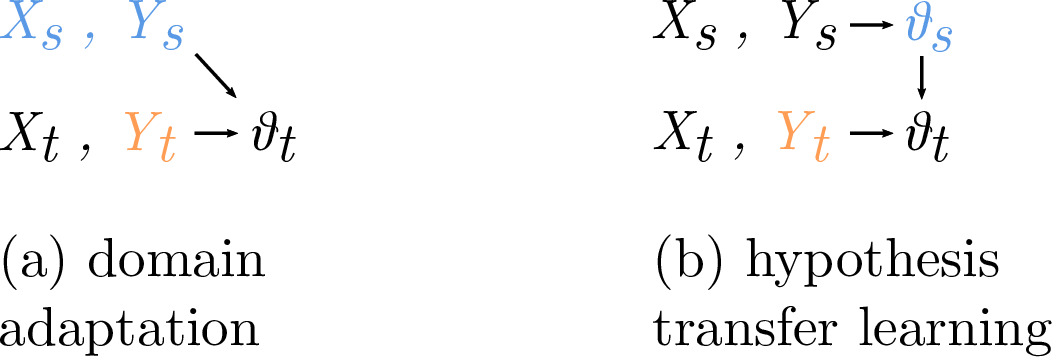}
\caption{Transfer learning uses \color[HTML]{5999E5}{accessible knowledge} \textcolor{black}{from a source domain to inform data-driven (a) or model-driven (b) learning on a target domain. In both cases, the unsupervised setting implies that} \color[HTML]{FF9D55}{target label data} \textcolor{black}{is unavailable.}}
\label{fig:transfer_learning}
\end{figure}

Commonly, a dataset distribution shift can occur due to diverse sources~\cite{quinonero2008dataset}: (i) domain shift, also known as covariate shift, is caused by hardware differences in data acquisition devices; (ii) feature distribution disparity is caused by population-level differences (e.g., gender, ethnicity) across domains; (iii) label distribution shift, where the proportional prevalence of labels in the source domain differs from that of the target domain. Due to the variety of distribution shifts, models have failed in real-world applications with shifted domains, thus posing an important threat to safety-critical applications.

Hypothesis transfer learning (HTL), also referred to as source-free domain adaptation (SFDA), addresses distribution shift under the non-transductive setting by using knowledge encoded in a model pretrained on the source domain to inform learning on the target domain (Fig.~\ref{fig:transfer_learning}b).
Unlike traditional domain adaptation (DA) approaches (Fig.~\ref{fig:transfer_learning}a), SFDA models do not have simultaneous access to the data from both source and target domains. This assumption mitigates the privacy and storage concerns arising in conventional DA methods (see Fig.~\ref{fig:transfer_learning} for the difference).


Extending ensemble learning to DA frameworks and, in particular, SFDA methods can uncover multiple modes within the source domain, improving the transferability of these models~\cite{lao2021hdmi}. However, the performance gain of an ensemble model is largely related to the diversity of its members. Particularly, averaging over identical networks or ensemble members with limited diversity is not better than a single model~\cite{rame2021dice}.

In this work, we encourage diversity among ensemble members in an unsupervised source-free domain adaptation setting where no labeled target data is available. While recent work in unsupervised SFDA has shown promising results, they either rely on a unique feature extractor~\cite{liang2020shot}, or one shared between an ensemble of source hypotheses~\cite{lao2021hdmi}, which leads to limited diversity in the function space of the source domain (see Sec.~\ref{sec:analysis} for analysis). 

Diversity in ensemble leads to the best-calibrated uncertainty estimators~\cite{lakshminarayanan2017simple}, and therefore the performance benefits of feature diversity within ensembles in out-of-distribution (OOD) settings~\cite{pagliardini2022agree}.
Other recent works in DNN analysis also show that different architectures tend to explore different representations~\cite{kornblith2019similarity,nguyen2021do,antoran2020depth,zaidi2021neural}.
Inspired by them, our work proposes to increase diversity by not only using separate feature extractors but also by introducing Distinct Backbone Architectures (\diversitycontrib) across hypotheses.

While a regularization approach to unconstrained mutual information (MI) maximization during adaptation is promising in low diversity settings~\cite{lao2021hdmi}, enforcing similarity between highly diverse hypotheses is insufficient to counteract the catastrophic impact of weak hypotheses when they inevitably arise as outliers. Therefore, we highlight the necessity of a trade-off between diversity and the amount of freedom each ensemble member can have. Hence, we introduce Penalized Diversity (\model), a new unsupervised SFDA approach that maximizes diversity exploitation via \diversitycontrib while mitigating the negative impact of Weak Hypotheses through the Penalization (\penaltycontrib) of their contribution by regularization.

In many real-world applications, the uniform distribution assumption between source and target does not hold. This assumption can negatively impact the performance of many current SFDA models under label distribution shift~\cite{liang2020shot, lao2021hdmi} (Sec. \ref{sec:experimental_results}, label shift experiments). We further extend PD to address the label distribution shift by introducing a weighted MI maximization based on estimation over target distribution.
Extensive experiments on multiple domain adaptation benchmarks (Office-31, Office-Home, and VisDA-C), medical, and digit datasets under covariate and label distribution shifts exhibit the effectiveness of \model.

\section{Approach}
\subsection{Preliminaries}
Assuming $\mathcal{X}$ indicates the input space and $\mathcal{Y}$ represents the output space, in an unsupervised source-free domain adaptation (uSFDA) setting, we denote the source domain as $\mathcal{D}_s = \{(x_i^S, y_i^S)\}_{i=1}^{N_s}$, where $x_i^S \in \mathcal{X^S}$ and $y_i^S \in \mathcal{Y^S}$. The unlabeled target domain is denoted as $\mathcal{D}_t = \{(x_i^T)\}_{i=1}^{N_t}$, where $x_i^T \in \mathcal{X^T}$ and $\mathcal{X}^S \neq \mathcal{X}^T$.
For now, we assume that the difference in the joint distribution $P(\mathcal{X}, \mathcal{Y})$ of source and target stems from the covariate shift only. Therefore, this induces a domain shift between the source and target domains, ($P_S(X) \neq P_T(X)$), whereas the learning task remains the same, with $\mathcal{Y^S} = \mathcal{Y^T}$ and $P_S(Y\mid X) = P_T(Y\mid X)$. Given a hypothesis space $\mathcal{H}$, uSFDA learns a source hypothesis $h_s: \mathcal{X}^S \xrightarrow{} \mathcal{Y}^S \in \mathcal{H}^S$ and a target hypothesis $h_t: \mathcal{X}^T \xrightarrow{} \mathcal{Y}^T \in \mathcal{H}^T$, to predict the unobserved target labels ${Y}_t^*$. From the Bayesian perspective, the predictive posterior distribution can be written as:
\begin{equation}\label{eq:posterior}
    \begin{split}
        &p({Y}_t^* \mid \mathcal{D}_s, \mathcal{D}_t) \\
        &= \int_{h_t} p({Y}_t^* \mid \mathcal{D}_t, h_t) \int_{h_s} p(h_t \mid \mathcal{D}_t, h_s) p(h_s \mid \mathcal{D}_s) dh_s dh_t\\
    \end{split}
\end{equation}
Eq.~\ref{eq:posterior} describes two learning phases; first, the posterior over the source hypothesis $p(h_s \mid \mathcal{D}_s)$ is learned using the source dataset $\mathcal{D}_s$, and second, the posterior over the target hypothesis $p(h_t \mid \mathcal{D}_t, h_s)$ is learned by marginalizing over samples of the source hypothesis adapted to the target domain, which only contains unlabeled examples.

\citet{liang2020shot} use a single model to estimate the distribution over the source hypothesis, and by extension the distribution over the target hypothesis. \citet{lao2021hdmi} improved this approximation by incorporating multiple hypotheses that {\it share} the same feature extraction backbone. While the latter is considered an ensemble, by definition, it is constrained by learning shared extracted features. In this paper, we promote diversity by introducing the use of Distinct Backbone Architectures (\diversitycontrib) across hypotheses. We argue and show empirically (Sec.~\ref{sec:experimental_results}) that this helps us achieve a better approximation of $p(h_t \mid \mathcal{D}_t, h_s)$ with higher diversity in the representation space.

However, unconstrained MI maximization during adaptation is prone to the induction of weak hypotheses due to error accumulation. The hypothesis disparity (HD) introduced by~\cite{lao2021hdmi} acts as a regularizer by enforcing similarity across hypotheses over the distribution of predicted labels. While this regularization showed promise in low diversity settings, enforcing similarity between highly diverse hypotheses is insufficient, and weak hypotheses inevitably arise (see Sec.~\ref{sec:analysis} for experiments). Unfortunately, the very nature of how similarity is computed in HD makes it highly vulnerable to weak hypotheses. We propose an approach that mitigates the negative impact of Weak Hypotheses through the Penalization (\penaltycontrib) of their contribution to the computation of HD when they arise as outliers (Sec.~\ref{sec:experimental_results}). 

In the following sections, we describe three main components of our proposed model, \modelname (\model). 

\subsection{Learning Diverse Source Hypotheses Using Distinct Backbone Architectures}\label{sec:dba}
To maximize the diversity of predictive features learned in the source domain, we propose removing any weight sharing between the backbones of separate hypotheses by introducing the use of distinct architectures. For example, on the LIDC dataset, our approach (\diversitycontrib) is implemented through the use of a mixture of ResNet10 and ResNet18 backbones.

We define the set of source hypotheses as $\{h_i^S : h_i^S = f_i^S \circ g_i^S \}_{i=1}^M$, where $\{f_i^S\}_{i=1}^M$ and $\{g_i^S \in \Psi^S\}_{i=1}^M$ represent the set of classifiers and the set of feature extractors, respectively, and M represents the number of hypotheses. We train each hypothesis using the cross entropy loss function ($CE$):
\begin{equation}\label{eq_source_loss}
 L_{S} = \argmin_{\mathcal{H^S}} \mathbb E_{(x,y) \in \mathcal{X}^S \times \mathcal{Y}^S}[CE(h(x), y)],~\forall h \in \{h_i^S\}_{i=1}^{M}
\end{equation}
\noindent where $h(x) = p(y\mid x; h)$ denotes the probability distribution of input $x$ predicted by hypothesis $h$ .

\subsection{Diversity Exploitation Through Weak Hypothesis Penalization}\label{sec:whp}
Assuming $\mathcal{D}_t = \{(x_i^T)\}_{i=1}^{N_t}$ is a set of unlabeled target samples, our goal is to effectively adapt the set of hypotheses trained on the source domain $\{h_i^S\}_{i=1}^M$ into a set of target hypotheses $\{h_i^T\}_{i=1}^M$. Due to the absence of both source data and labeled target data during the adaptation phase, we maximize the mutual information (MI) between the target data distribution ($X^T$) and the predictions by the target hypotheses ($\hat{Y}^T$)~\cite{liang2020shot} using Eq.~\ref{eq:mi}.
\begin{equation}\label{eq:mi}
    \max_{\Psi^T} \mathbb E_{x \in \mathcal{X}^T} [I(X^T; h(X^T))],~\forall h \in \{h_i^T\}_{i=1}^{M}
\end{equation}
\noindent where $I(X^T; \hat{Y}^T) = H(\hat{Y}^T) - H(\hat{Y}^T \mid X^T)$ is defined as MI with $H$ indicating entropy, $\hat{Y}^T$ is the predicted output of $h(X^T)$, and $\Psi^T$ is the space of target feature extractors. Assuming that only covariate shift is present, both the source and the target domains share the same label space, so we keep the parameters for the classifiers $f_i^T$ fixed while updating the feature extractors $g_i^T \in \Psi^T$.

Unconstrained unsupervised training of target hypothesis ensembles solely using MI maximization results in undesirable target label prediction disagreements. We use the hypothesis disparity (HD) regularization to marginalize out these disagreements~\cite{lao2021hdmi}. HD measures the dissimilarity between the predicted label probability distributions among pairs of hypotheses over the input space $\mathcal{X}$:
\begin{equation}\label{eq:original_HD}
    \text{HD}_{h_i, h_j \in \{h^T\}, i \neq j} (h_i, h_j) = \int_{\mathcal{X}} d(h_i(x), h_j(x)) p(x) dx
\end{equation}
\noindent where $d(.)$ defines the dissimilarity metric. Throughout this study, we use cross entropy to measure dissimilarity.

In its original formulation, computing HD relies on randomly selecting a single hypothesis that serves as an anchor (reference) for the pairwise disparity measures with the rest of the $M\!\!-\!\!1$ hypotheses. This selected anchor remains fixed throughout the training process. We note that this method of choosing the anchor may potentially have a catastrophic impact; a weak performance hypothesis chosen as the anchor could act as an attractor and collapse the model (See~\ref{seq:anchor_selection} and Table~\ref{tab:anchor_comparison_office_31}). 

In order to address this issue, we redefine HD (Eq.~\ref{eq:original_HD}) by proposing Weak Hypothesis mitigation through Penalization (\penaltycontrib), which constructs the anchor hypothesis (\(h_{\penaltycontriblower}\)) as an ensemble where the contribution of each hypothesis is weighted according to its cosine similarity to other hypotheses (Eq.~\ref{eq:weighted_ensemble} to Eq.~\ref{eq:redefine_HD}):
\begin{equation}\label{eq:weighted_ensemble}
    h_{\penaltycontriblower}(X^T) = \sum_{i=1}^{M} \hat{w}_i h_i(X^T),
\end{equation}
\noindent where $h_i \in \{h^T\}$, and $\hat{w}_i$ represents the normalized weight for each hypothesis and is computed as:
\begin{equation}\label{eq:normalized_weighted_ensemble}
    \hat{w}_i = \frac{exp(w_i)}{\sum_{{j}=1}^{M} exp(w_{{j}})}
\end{equation}
\begin{equation}\label{eq:cosine_similarity}
   w_i = \mathbb E_{h_{j} \in \{h^T\}, j \neq i } [\frac{h_i(X^T)\cdot h_j(X^T)}{\vert h_i(X^T) \vert \cdot \vert h_j(X^T)\vert}], ~  \forall i.
\end{equation}
\begin{equation}\label{eq:redefine_HD}
    \text{HD}_{h_i \in \{h^T\}} (h_i, h_{\penaltycontriblower}) = \int_{\mathcal{X}} d(h_i(x), h_{\penaltycontriblower}(x)) p(x) dx
\end{equation}
\noindent In effect, the contribution to the ensemble anchor \(h_{\penaltycontriblower}\) of a more distant hypothesis $h_i$, based on its marginal cosine similarity to other hypotheses, is penalized through the reduction of $w_i$. Hence, we improve performance by diminishing the probability of selecting a weak anchor. In Section~\ref{seq:anchor_selection}, we compare \penaltycontrib to alternative strategies for anchor hypothesis selection and show its superior performance.

\subsection{Target Distribution Estimation Via Pseudo-Labels}
We take one step further and refine our assumption on possible shifts in the joint distribution of source and target. Instead of assuming that the changes only stemmed from the difference in their marginal ($P(X)$), we also allow shifts in their prior, i.e. $P_S(Y) \neq P_T(Y)$, namely label distribution shift. 
For \model to be able to perform under label distribution shift, we suggested weighted mutual information (MI) based on the proportion of target classes. Since the exact proportion of target classes is not accessible in an unsupervised setting, an estimation using pseudo-labeling~\citep{liang2020shot} can be used. 
The label entropy for MI maximization is reweighted according to the estimated class proportions and reformulates the MI maximization as:
\begin{equation}
    I_{W}(X^T; \hat{Y}^T) = W \ast H(\hat{Y}^T) - H(\hat{Y}^T \mid X^T)
\end{equation}
\noindent where $W=[w_1,\dots,w_C]$ is an estimated class proportion from pseudo-labels, and $w_i = \frac{n_{c_i}}{\sum_{j=1}^{C}n_{c_j}}$, where $n_{c_i}$ represents the number of samples in class $c_i$ and $C$ is the total number of classes.\\

In summary, our full objective for target training is a combination of weighted mutual information and hypothesis disparity regularization.
\begin{equation}\label{eq:pd}
    L_T = \alpha \mathbb E_{h \in \mathcal{H}^T} [-I_{W}(X^T; h(X^T))] + \beta \mathbb E_{h \in \{h^T\}} [\text{HD}(h, h_{\penaltycontriblower})]
\end{equation}
\noindent where $\alpha$ and $\beta$ are hyperparameters indicating the contribution of each of MI and HD in the target training.

\section{Experiments and Results}

\subsection{Datasets}
To validate our model under covariate shift, we consider natural and medical image datasets. For the natural images, we consider domain adaptation benchmark datasets, namely \textbf{Office-31}~\cite{saenko2010office_31}, \textbf{Office-Home}~\cite{venkateswara2017deep}, and \textbf{VisDA-C}~\cite{peng2018visda}. For the medical application, we evaluate our model on the \textbf{LIDC}~\cite{armato2011lidc} dataset.
Office-31 dataset includes three domains that share a set of 31 classes; Amazon (A), DSLR (D), and Webcam (W). Office-Home has four domains, each having 65 classes; Artistic images (AR), Clip art (CL), Product images (PR), and Real-World images (RW). VisDA-C has 12 classes with synthetic images in the source domain and real images in the target domain.
For our medical imaging experiment, we divided the LIDC dataset into four domains based on the manufacturer of the data-capturing device: GE\_medical (G), Philips (P), SIEMENS (S), and TOSHIBA (T). Each of these domains has two classes, healthy and unhealthy. We validate the existence of covariate shift across LIDC domains, from both statistical and experimental perspectives. A detailed statistical analysis is provided in the Appendix~\ref{sec:A1}. For the experiments on label shift, we consider synthetic digit datasets--\textbf{MNIST} (M)~\cite{lecun1998mnist}, \textbf{MNIST-M} (N)~\cite{ganin2016mnist_m}, and \textbf{USPS} (U)~\cite{hull1994usps}.

\subsection{Baselines}
Unsupervised transfer learning approaches can be categorized as either unsupervised domain adaptation (UDA) or SFDA, depending on whether or not they require access to the source data during the adaptation phase. We consider baselines from both sets. For UDA, we compare \model to DANN~\cite{ganin2015unsupervised}, DAN~\cite{long2015learning}, CDAN~\cite{long2018conditional}, SAFN+ENT~\cite{xu2019larger}, rRevGrad+CAT~\cite{deng2019cluster},  MDD~\cite{zhang2019bridging}, and MCC~\cite{jin2020minimum}. For SFDA, we use AdaBN~\cite{li2016adabn}, Tent~\cite{wang2021tent}, SHOT~\cite{liang2020shot}, HDMI~\cite{lao2021hdmi}, and NRC~\cite{yang2021nrc}. We also consider the performance of source hypotheses at directly predicting target labels as a Source-only model, and MI-ensemble as a model with three hypotheses with only MI maximization and no regularizer.

\begin{table*}[!htb]
\centering
\caption{\small{Target accuracy ($\%$) on Office-31 under covariate shift (source $\rightarrow$ target). In this and all the following tables, $^\dag$ represents results reported from our implementations.}}\label{tab:results_office_31}
\begin{adjustbox}{width=0.85\textwidth}
\begin{tabular}{lcccccccc}
\toprule
Method & Source-free & A$\rightarrow$D & A$\rightarrow$W & D$\rightarrow$A & D$\rightarrow$W & W$\rightarrow$A & W$\rightarrow$D & Avg. \\
\midrule
Source-only$^\dag$   & \xmark & 78.6 & 80.5 & 63.6 & 97.1 & 62.8 & 99.6 & 80.4 \\
DAN~\cite{long2015learning}           & \xmark & 78.6 & 80.5 & 63.6 & 97.1 & 62.8 & 99.6 & 80.4 \\
DANN~\cite{ganin2015unsupervised}          & \xmark & 79.7 & 82.0 & 68.2 & 96.9 & 67.4 & 99.1 & 82.2 \\
SAFN+ENT~\cite{xu2019larger}      & \xmark & 90.7 & 90.1 & 73.0 & 98.6 & 70.2 & 99.8 & 87.1 \\
rRevGrad+CAT~\cite{deng2019cluster}   & \xmark & 90.8 & 94.4 & 72.2 & 98.0 & 70.2 & \textbf{100.} & 87.6 \\
MDD~\cite{zhang2019bridging}           & \xmark & 93.5 & \textbf{94.5} & 74.6 & 98.4 & 72.2 & \textbf{100.} & 88.9 \\
MCC~\cite{jin2020minimum} & \xmark & 95.5 & 98.6 & 100 & 94.4 & 72.9 & 74.9 & 89.4 \\
\midrule
MI-ensemble$^\dag$         & \cmark & 91.0 & 93.0 & 72.3 & 96.5 & 73.7 & 97.4 & 87.3 \\
AdaBN~\cite{li2016adabn} & \cmark & 81.0 & 82.4 & 67.2 & 97.7 & 68.2 & 99.8 & 82.7 \\
Tent~\cite{wang2021tent} & \cmark & 82.1 & 85.1 & 68.8 & 97.5 & 63.0 & 99.8 & 82.7 \\
SHOT~\cite{liang2020shot}          & \cmark & 93.1 & 90.9 & 74.5 & 98.8 & 74.8 & 99.9 & 88.7 \\
HDMI~\cite{lao2021hdmi} & \cmark & 94.4 & 94.0 & 73.7 & 98.9 & 75.9 & 99.8 & 89.5 \\
NRC~\cite{yang2021nrc} & \cmark & \textbf{96.0} & 90.8 & \textbf{75.3} & \textbf{99.0} & 75.0 & \textbf{100.} & 89.4 \\
\midrule
\model & \cmark & 95.6 & 94.3 & \textbf{75.3} & 98.7 & \textbf{76.4} & 99.8 & \textbf{90.0} \\

\bottomrule
\end{tabular}
\end{adjustbox}
\end{table*}

\begin{table*}[!htb]
\centering
\caption{\small{Target accuracy ($\%$) on Office-Home under covariate shift (source $\rightarrow$ target).}}\label{tab:results_office_home}
\begin{adjustbox}{width=1\textwidth}
\begin{tabular}{lcccccccccccccc}
\toprule
Method & Source-free & Ar$\rightarrow$Cl & Ar$\rightarrow$Pr & Ar$\rightarrow$Rw & Cl$\rightarrow$Ar & Cl$\rightarrow$Pr & Cl$\rightarrow$Rw & Pr$\rightarrow$Ar & Pr$\rightarrow$Cl & Pr$\rightarrow$Rw & Rw$\rightarrow$Ar & Rw$\rightarrow$Cl & Rw$\rightarrow$Pr & Avg. \\
\midrule
Source-only$^\dag$   & \xmark & 45.6 & 69.2 & 76.5 & 55.3 & 64.4 & 67.4 & 55.1 & 41.6 & 74.4 & 66.0 & 46.3 & 79.4 & 61.8 \\
DAN~\cite{long2015learning}           & \xmark & 43.6 & 57.0 & 67.9 & 45.8 & 56.5 & 60.4 & 44.0 & 43.6 & 67.7 & 63.1 & 51.5 & 74.3 & 56.3 \\
DANN~\cite{ganin2015unsupervised}          & \xmark & 45.6 & 59.3 & 70.1 & 47.0 & 58.5 & 60.9 & 46.1 & 43.7 & 68.5 & 63.2 & 51.8 & 76.8 & 57.6 \\
SAFN~\cite{xu2019larger}          & \xmark & 52.0 & 71.7 & 76.3 & 64.2 & 69.9 & 71.9 & 63.7 & 51.4 & 77.1 & 70.9 & 57.1 & 81.5 & 67.3 \\
MDD~\cite{zhang2019bridging}           & \xmark & 54.9 & 73.7 & 77.8 & 60.0 & 71.4 & 71.8 & 61.2 & 53.6 & 78.1 & 72.5 & \textbf{60.2} & 82.3 & 68.1 \\
\midrule
MI-ensemble$^\dag$           & \cmark & 55.2 & 71.9 & 80.2 & 62.6 & 76.8 & 77.8 & 63.2 & 53.8 & 81.1 & 67.9 & 58.3 & 81.4 & 69.2 \\
AdaBN~\cite{li2016adabn} & \cmark & 50.9 & 63.1 & 72.3 & 53.2 & 62.0 & 63.4 & 52.2 & 49.8 & 71.5 & 66.1 & 56.1 & 77.1 & 61.5 \\
Tent~\cite{wang2021tent} & \cmark & 47.9 & 66.0 & 73.3 & 58.8 & 65.9 & 68.1 & 60.2 & 47.3 & 75.4 & 70.8 & 54.0 & 78.7 & 63.9 \\
SHOT~\cite{liang2020shot}           & \cmark & 56.9 & 78.1 & 81.0 & 67.9 & 78.4 & 78.1 & 67.0 & 54.6 & 81.8 & 73.4 & 58.1 & \textbf{84.5} & 71.6 \\
HDMI~\cite{lao2021hdmi}          & \cmark & 57.8 & 76.7 & 81.9 & 67.1 & 78.8 & \textbf{78.8} & 66.6 & 55.5 & 82.4 & 73.6 & 59.7 & 84.0 & 71.9 \\
NRC~\cite{yang2021nrc} & \cmark & 57.7 & \textbf{80.3} & \textbf{82.0} & 68.1 & \textbf{79.8} & 78.6 & 65.3 & 56.4 & 83.0 & 71.0 & 58.6 & 85.6 & 72.2 \\
\midrule
\model          & \cmark & \textbf{58.9} & 78.0 & 80.9 & \textbf{69.3} & 76.7 & 76.9 & \textbf{69.6} & \textbf{56.5} & \textbf{83.4} & \textbf{75.1} & 59.9 & \textbf{84.5} & \textbf{72.5} \\

\bottomrule
\end{tabular}
\end{adjustbox}
\end{table*}

\begin{table*}[!htb]
\centering
\caption{\small{Target accuracy ($\%$) on LIDC under covariate shift (source $\rightarrow$ target). The results are averaged over five different runs.}}\label{tab:results_lidc}
\begin{adjustbox}{width=1\textwidth}
\begin{tabular}{lcccccccccccccc}
\toprule
Method & Source-free & G$\rightarrow$P & G$\rightarrow$S & G$\rightarrow$T & P$\rightarrow$G & P$\rightarrow$S & P$\rightarrow$T & S$\rightarrow$G & S$\rightarrow$P & S$\rightarrow$T & T$\rightarrow$G & T$\rightarrow$P & T$\rightarrow$S & Avg. \\
\midrule
Source-only$^\dag$ & \xmark & 65.6 & 65.9 & 45.9 & 61.9 & 60.5 & 49.3 & 65.2 & 66.9 & 57.1 & 50.0 & 50.0 & 50.0 & 57.2 \\ 

DAN~\cite{long2015learning}$^\dag$ & \xmark & 59.3 & 65.8 &	33.9 &	61.1 & 59.7 & 30.7 & 56.9 &	58.6 &	40.2 &	48.7 &	45.9 &	47.6 & 50.7 \\

MDD~\cite{zhang2019bridging}$^\dag$ & \xmark & 64.1 & 63.6 & 57.7 & 54.7 & 59.8 & 47.7 & \textbf{67.2} & 63.4 & 59.1 & 50.0 &	50.3 &	50.0 & 57.3 \\

\midrule
MI-ensemble$^\dag$ & \cmark & 67.1 & 63.4 & \textbf{66.6} & 62.9 & 63.2 & 52.1 & 65.1 & 64.9 & 55.2 & 60.8 & 58.6 & 58.2 & 61.5  \\

SHOT~\cite{liang2020shot}$^\dag$ & \cmark & 67.0 & \textbf{67.1} & 61.6 & 59.9 & 56.9 & 53.2 & 66.8 & \textbf{69.0} & 66.1 & 60.4 & 61.0 & 54.9 & 61.9 \\

HDMI~\cite{lao2021hdmi}$^\dag$ & \cmark & 67.1 & 66.6 & 64.6 & 65.4 & 64.2 & 54.6 & 66.2 & 65.1 & 54.6 & 60.7 & \textbf{60.0} & 58.7 & 62.3 \\

\midrule

\model & \cmark & \textbf{68.9} & 65.9 & 60.9 & \textbf{65.6} & \textbf{65.6} & \textbf{54.8} & 65.8 & 66.9 & \textbf{66.6} & \textbf{61.2} & 59.6 & \textbf{59.6} & \textbf{63.5} \\

\bottomrule
\end{tabular}
\end{adjustbox}
\end{table*}
For label distribution shift experiments, aside from comparing with two SFDA models namely SHOT and HDMI, we compare \model with MARS~\cite{rakotomamonjy2022mars} and OSTAR~\cite{kirchmeyer2022ostar}, the two recent state-of-the-art models for label distribution shift. MARS~\cite{rakotomamonjy2022mars} proposed based on two estimating proportion strategies, where hierarchical clustering defines MARSc and Gaussian mixtures indicates MARSg. In nearly all our experiments, MARSc outperforms MARSg, therefore we only report the performance of MARSc while referring to it as MARS.

\subsection{Experimental Setup}\label{sec:experimental_setup}
In the experiments presented in this paper, we consider both covariate shift and label distribution shift between source and target domains. We simply instill diversity in DBA using different depths of a given architecture~\cite{antoran2020depth,zaidi2021neural}. The code base of \model is built upon SHOT. Further details on the hyperparameters are provided in the Appendix~\ref{sec:A1}.

\subsection{Results}\label{sec:experimental_results}
In this section, we present the performance of our model in comparison with the baselines in each benchmark dataset under different distributional shifts.

\begin{table}
\centering
\caption{\small{Target accuracy ($\%$) on VisDA-C under covariate shift.}}\label{tab:results_visda}
\begin{adjustbox}{width=0.5\textwidth}
\begin{tabular}{lcc}
\toprule
Method & Source-free & Avg. per-class accuracy \\
\midrule
Source-only$^\dag$   & \xmark & 44.6 \\
DAN~\cite{long2015learning} & \xmark & 61.1 \\
CDAN~\cite{long2018conditional}   & \xmark & 70.0 \\
MDD~\cite{zhang2019bridging}  & \xmark & 74.6 \\
MCC~\cite{jin2020minimum} & \xmark & 78.8 \\
\midrule
Tent~\cite{wang2021tent} & \cmark & 65.7 \\
SHOT~\cite{liang2020shot} & \cmark & 79.6 \\
HDMI~\cite{lao2021hdmi}  & \cmark & 82.4 \\
\midrule
\model  & \cmark & \textbf{83.8} \\
\bottomrule
\end{tabular}
\end{adjustbox}
\end{table}
\subsubsection{Natural Images}
The performance of our model alongside the baselines on Office-31, Office-Home, and VisDA-C under covariate shift are presented in Table~\ref{tab:results_office_31}, ~\ref{tab:results_office_home}, and~\ref{tab:results_visda} respectively, where it can be observed that our \modelname (\model) approach outperforms all UDA and SFDA baselines.

\subsubsection{Medical Dataset}\label{sec:lidc_result}
The experimental results on the LIDC dataset, given four different domains, are depicted in Table~\ref{tab:results_lidc}. The results indicate the effectiveness of PD in comparison with other baselines. Using \diversitycontrib with \penaltycontrib increases the performance from $62.3\%$ to $63.5\%$. 

\subsubsection{Digit Dataset}
The effect of using weighted label entropy in our modified MI maximization objective in the experiments on digit datasets is presented in Table~\ref{tab:digit_label_shift}. Following \citep{azizzadenesheli2019regularized}, we used two strategies, namely \textit{Tweak-One shift} and \textit{Minority-Class shift} with a probability value of $p$ to create a label distribution shift in each of the datasets. In \textit{Tweak-One shift} ($L_t$), one of the classes is randomly selected whereas, in \textit{Minority-Class shift} ($L_m$), a subset of classes (in our experiments, 5 out of 10 classes) is chosen randomly. Then the proportion of the chosen class(es) in the target domain changes by value $p$, i.e. keeping only $p\%$ of the samples in the chosen class (see Fig.~\ref{fig:sensitivity_data_dist} (b) and (c) for the distribution of labels in each dataset). As demonstrated in Table~\ref{tab:digit_label_shift}, using an estimation of target label distribution as weights in MI objective mitigates the impact of label shift. The improvement is more notable in $L_m$ experiment (more than $8\%$ improvement over OSTAR). In these experiments, covariate shift is also present.
\begin{table*}[!htb]
\centering
\caption{\small{Target accuracy ($\%$) on digit datasets with no label shift ($N_l$), tweak-one ($L_t$), and minority-class ($L_m$) label distribution shift with $p=0.1$ (source $\rightarrow$ target). 
}}\label{tab:digit_label_shift}
\begin{adjustbox}{width=0.9\textwidth}
\begin{tabular}{lccccccccc}
\toprule
Method & Strategy & Source-free & M$\rightarrow$U & M$\rightarrow$N & U$\rightarrow$M & U$\rightarrow$N & N$\rightarrow$M & N$\rightarrow$U &  Avg. \\
\midrule
OSTAR~\cite{kirchmeyer2022ostar}$^\dag$ & & \xmark & 96.5 & 33.2 & \textbf{98.4} & 34.0 & \textbf{98.4} & 90.3 & 75.1 \\
MARS~\cite{rakotomamonjy2022mars}$^\dag$ & & \xmark & 92.3 & 40.1 & 96.8 & \textbf{38.6} & 95.4 & 88.2 & 75.2 \\
SHOT~\cite{liang2020shot}$^\dag$ & $N_l$ & \cmark & 88.9 & 45.9 & 93.2 & 29.7 & 95.4 & 88.5 & 73.6 \\
HDMI~\cite{lao2021hdmi}$^\dag$ &  & \cmark & 95.2 & 49.6 & 95.0 & 26.5 & 96.2 & 93.1 & 75.9 \\
\model & & \cmark & \textbf{96.9} & \textbf{50.1} & 95.6 & 26.4 & 96.6 & \textbf{97.6} & \textbf{77.2} \\
\midrule
OSTAR~\cite{kirchmeyer2022ostar}$^\dag$ & & \xmark & 94.6 & 37.9 & \textbf{98.3} & 27.1 & \textbf{97.4} & 85.6 & 73.5 \\
MARS~\cite{rakotomamonjy2022mars}$^\dag$ & & \xmark & 97.4 & 44.4 & 96.2 & \textbf{44.6} & 90.9 & 89.0 & \textbf{77.0} \\
SHOT~\cite{liang2020shot}$^\dag$ & $L_t$ & \cmark & 84.5 & 46.2 & 89.3 & 30.6 & 89.5 & 83.6 & 70.6 \\
HDMI~\cite{lao2021hdmi}$^\dag$ &  & \cmark & 89.9 & 48.7 & 89.6 & 27.3 & 90.5 & 87.2 & 72.2 \\
\model & & \cmark & \textbf{97.6} & \textbf{49.1} & 92.9 & 29.1 & 96.4 & \textbf{97.0} & \textbf{77.0} \\
\midrule
OSTAR~\cite{kirchmeyer2022ostar}$^\dag$ & & \xmark & 58.6 & 37.1 & \textbf{96.5} & 23.1 & 84.1 & 67.4 & 61.1 \\
MARS~\cite{rakotomamonjy2022mars}$^\dag$ & & \xmark & 59.9 & 23.3 & 92.6 & \textbf{35.3} & 80.0 & 69.3 & 60.1 \\
SHOT~\cite{liang2020shot}$^\dag$ & $L_m$ & \cmark & 57.1 & 43.9 & 58.2 & 28.7 & 60.9 & 56.7 & 50.9 \\
HDMI~\cite{lao2021hdmi}$^\dag$ &  & \cmark & 62.4 & 46.0 & 60.1 & 25.2 & 62.6 & 58.9 & 52.5 \\
\model & & \cmark & \textbf{87.5} & \textbf{47.7} & 84.3 & 31.8 & \textbf{85.0} & \textbf{78.6} & \textbf{69.2} \\
\bottomrule
\end{tabular}
\end{adjustbox}
\end{table*}
\begin{table*}[!htb]
\centering
\caption{\small{Ablation study on anchor selection under covariate shift: target accuracy ($\%$) on LIDC dataset with \diversitycontrib under different anchor selection strategies (source $\rightarrow$ target). 3H represents models with 3 hypotheses. The results are averaged over five different runs.}}\label{tab:anchor_comparison_lidc}
\begin{adjustbox}{width=1\textwidth}
\begin{tabular}{lccccccccccccc}
\toprule
Method & G$\rightarrow$P & G$\rightarrow$S & G$\rightarrow$T & P$\rightarrow$G & P$\rightarrow$S & P$\rightarrow$T & S$\rightarrow$G & S$\rightarrow$P & S$\rightarrow$T & T$\rightarrow$G & T$\rightarrow$P & T$\rightarrow$S &  Avg. \\
\midrule

3H-Fixed & 68.1 & \textbf{66.5} & 60.7 & 64.5 & 64.2 & 55.5 & 65.4 & \textbf{67.8} & 58.6 & 59.7 & 58.5 & 59.1 & 62.4 \\

3H-Random & \textbf{69.3} & 65.7 & \textbf{64.5} & 64.9 & 64.3 & \textbf{55.6} & \textbf{66.4} & 66.8 & 57.3 & 60.4 & 58.5 & \textbf{59.9} & 62.8 \\

3H-Ensemble & 69.2 & 66.2 & 59.3 & 65.3 & 64.7 & 54.2 & 65.9 & 67.1 & 56.9 & 61.0 & 59.9 & \textbf{59.9} & 62.6 \\

3H-\penaltycontrib & 68.9 & 65.9 & 60.9 & \textbf{65.6} & \textbf{65.6} & 54.8 & 65.8 & 66.9 & \textbf{66.6} & \textbf{61.2} & \textbf{59.6} & 59.6 & \textbf{63.5} \\

\bottomrule
\end{tabular}
\end{adjustbox}
\end{table*}
\begin{table}[t]
\centering
\caption{\small{Ablation study on anchor selection under covariate shift: target accuracy ($\%$) on Office-31 dataset with \diversitycontrib (3 hypotheses) under different anchor selection strategies (source $\rightarrow$ target).}}\label{tab:anchor_comparison_office_31}
\begin{adjustbox}{width=0.5\textwidth}
\begin{tabular}{lccccccc}
\toprule
Method & A$\rightarrow$D & A$\rightarrow$W & D$\rightarrow$A & D$\rightarrow$W & W$\rightarrow$A & W$\rightarrow$D & Avg. \\
\midrule

Fixed & 94.2 & 93.8 & 71.1 & 98.5 & 71.0 & \textbf{99.8} & 88.1  \\

Random & 94.0 & 94.1 & 70.3 & 98.5 & 70.4 & \textbf{99.8} & 87.9 \\

Ensemble & 95.4 & 94.2 & 71.5 & 98.5 & 71.1 & \textbf{99.8} & 88.4  \\

\penaltycontrib & \textbf{95.6} & \textbf{94.3} & \textbf{75.3} & \textbf{98.7} & \textbf{76.4} & \textbf{99.8} & \textbf{90.0} \\

\bottomrule
\end{tabular}
\end{adjustbox}
\end{table}

\subsection{Analysis}\label{sec:analysis}
\subsubsection{\diversitycontrib increases the diversity of source hypotheses}
\begin{figure*}[t] 
\centering
  \raisebox{\dimexpr 1.6cm}{\rotatebox{90}{G $\rightarrow$ P}}
  \centering
  \addtocounter{subfigure}{-1}
  \subfloat{\includegraphics[scale=0.51]{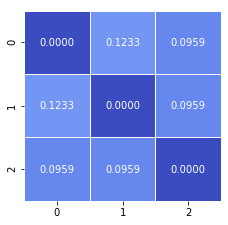}}
  \centering
  \addtocounter{subfigure}{-1}
  \subfloat{\includegraphics[scale=0.51]{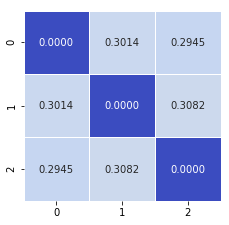}}
  \centering
  \addtocounter{subfigure}{-1}
  \subfloat{\includegraphics[scale=0.51]{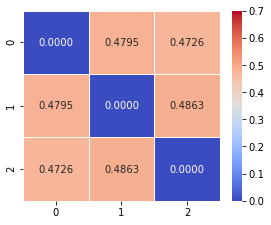}}
  \\
  \raisebox{\dimexpr 1.6cm}{\rotatebox{90}{S $\rightarrow$ T}}
  \centering
  \addtocounter{subfigure}{-1}
  \subfloat{\includegraphics[scale=0.51]{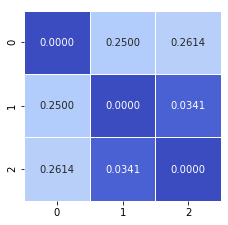}}
  \centering
  \addtocounter{subfigure}{-1}
  \subfloat{\includegraphics[scale=0.51]{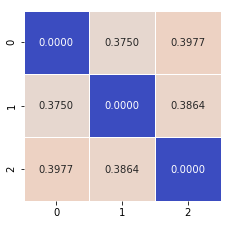}}
  \centering
  \addtocounter{subfigure}{-1}
  \subfloat{\includegraphics[scale=0.51]{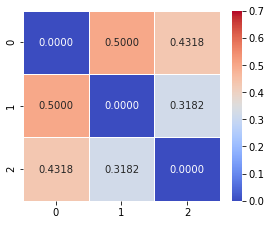}}
  \\
  \raisebox{\dimexpr 1.6cm}{\rotatebox{90}{A $\rightarrow$ D}}
  \centering
  \subfloat[\scriptsize{\sharedb} ]{\includegraphics[scale=0.513]{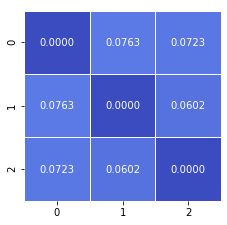}}
  \centering
  \subfloat[\scriptsize{\separateb} ]{\includegraphics[scale=0.513]{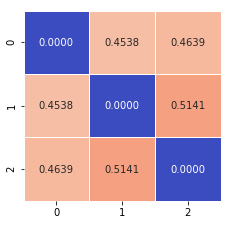}}
  \centering
  \subfloat[\scriptsize{\diversitycontrib}]{\includegraphics[scale=0.513]{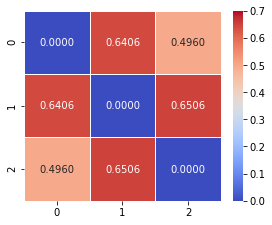}}
  \caption{\small{{\bf Diversity of the source hypotheses} based on the choice of feature extraction backbone on G $\rightarrow$ P (first row) and S $\rightarrow$ T (second row) from LIDC, and A $\rightarrow$ D from Office-31 datasets using disagreement of predictions between three hypotheses. \textit{Left plot}: three hypotheses with shared features extractors (\sharedb). \textit{Middle plot}: three hypotheses with separate feature extractors (no weight sharing) (\separateb). \textit{Right plot}: three hypotheses with distinct backbone (\diversitycontrib).
  }} \label{fig:diversity}
\end{figure*}

In order to investigate the relative impact on the diversity of introducing separate source hypothesis backbones with the same architecture and using distinct backbone architectures, we follow~\cite{fort2019deep} and measure the source hypotheses' disagreement in function space. More specifically, given a set of target samples $X$, we compute $\frac{1}{N} \sum_{i=1}^{M} \sum_{j=1}^{M} [f(X; \theta_i) \neq f(X; \theta_j)]$, where $N$ is the total number of target samples, $M$ defines the number of hypotheses, and $f(.)$ indicates the predicted class. Note that in this experiment, we analyze the diversity of the source hypotheses and no adaptation to the target dataset is made. We consider three ways of constructing the ensemble: 1) shared feature extractors, referred to as Shared Backbone (\sharedb), 2) hypotheses that are given separate feature extractors with the same backbone architecture, referred to as Separate Backbone (\separateb), and 3) \diversitycontrib (used in \model), which has separate feature extractors with distinct backbones.
Figure~\ref{fig:diversity} shows that 
simply introducing separate feature extractors for each hypothesis (\separateb) leads to a marked increase in diversity compared to sharing a feature extractor (\sharedb), as in HDMI. However, the largest diversity increase comes from the introduction of \diversitycontrib.

\begin{figure*}[!htb] 
  \centering
  \subfloat[\sharedb]{\includegraphics[scale=0.2]{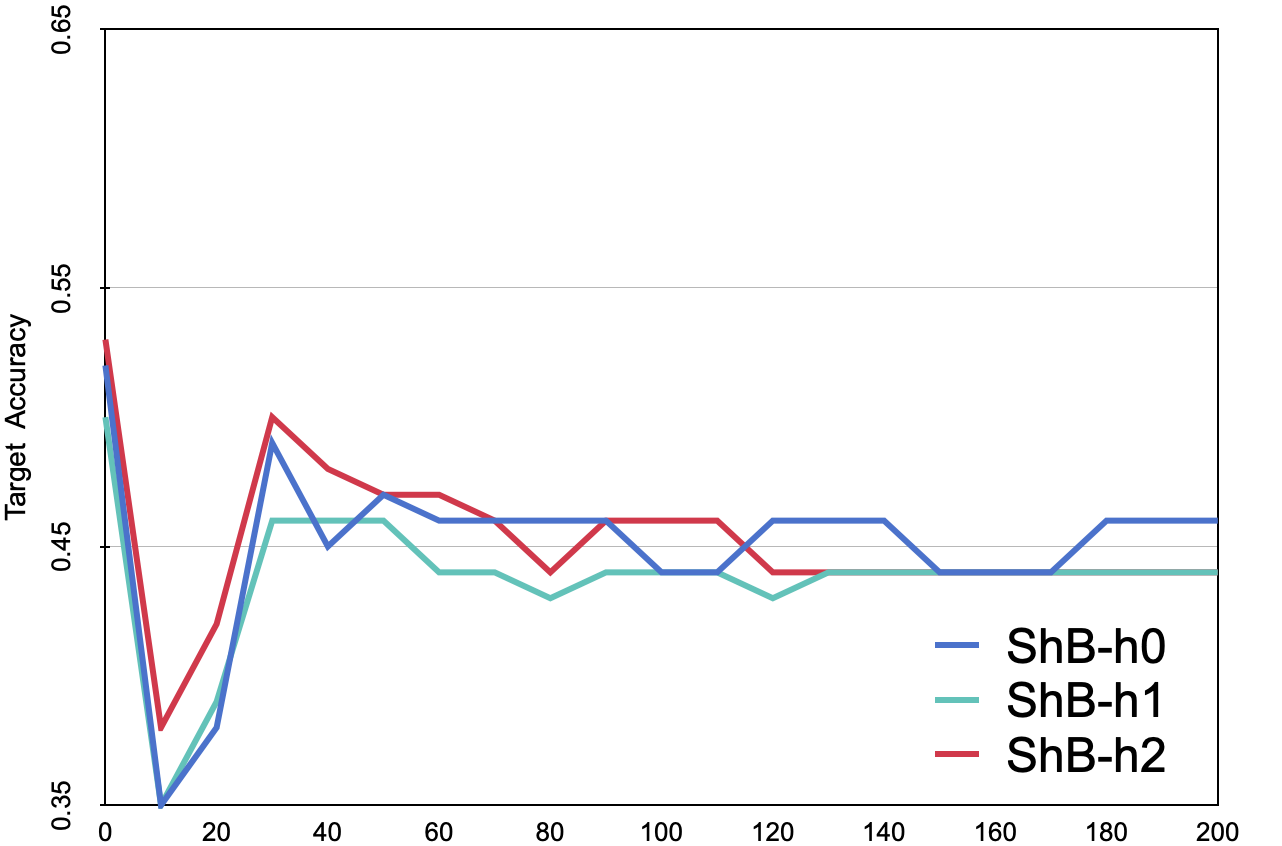}}
  \subfloat[\separateb]{\includegraphics[scale=0.2]{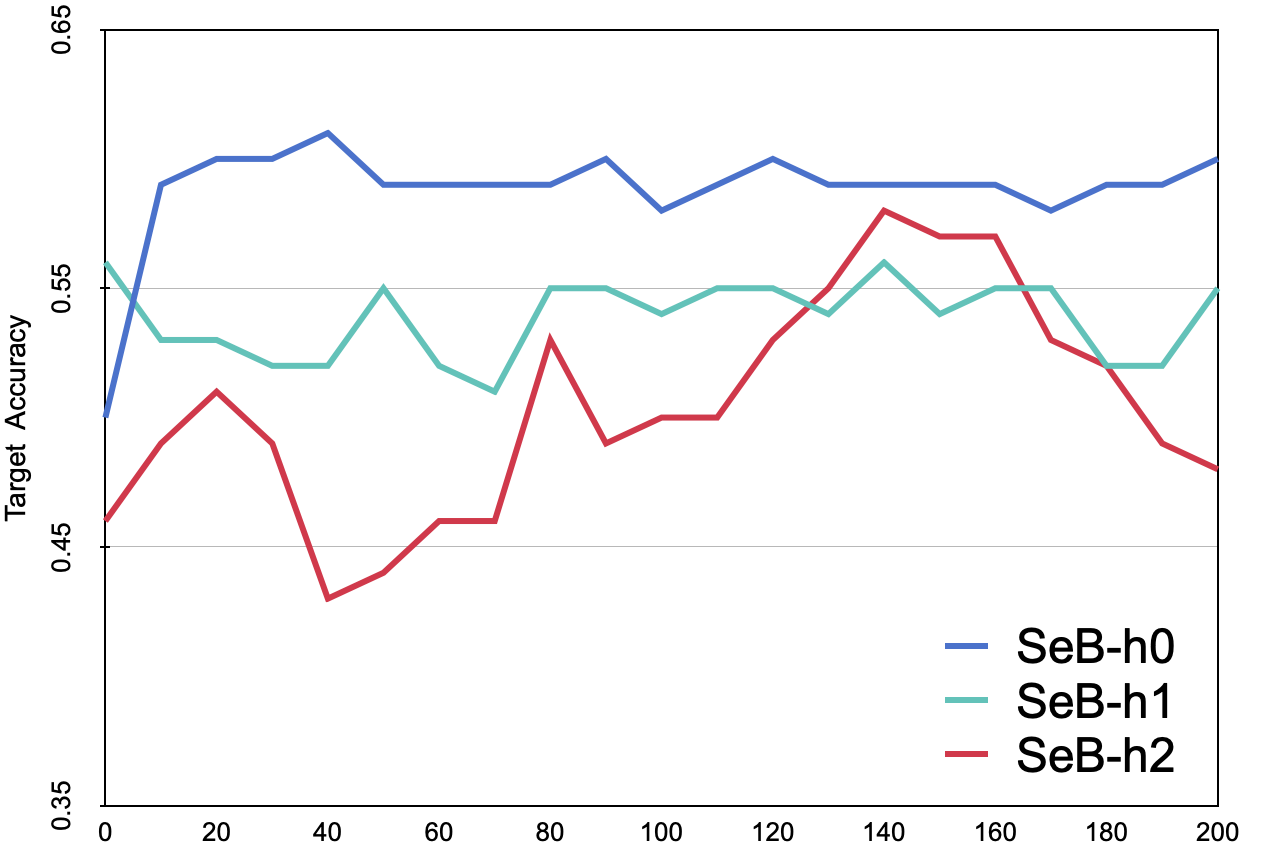}}
  \subfloat[\diversitycontrib]{\includegraphics[scale=0.2]{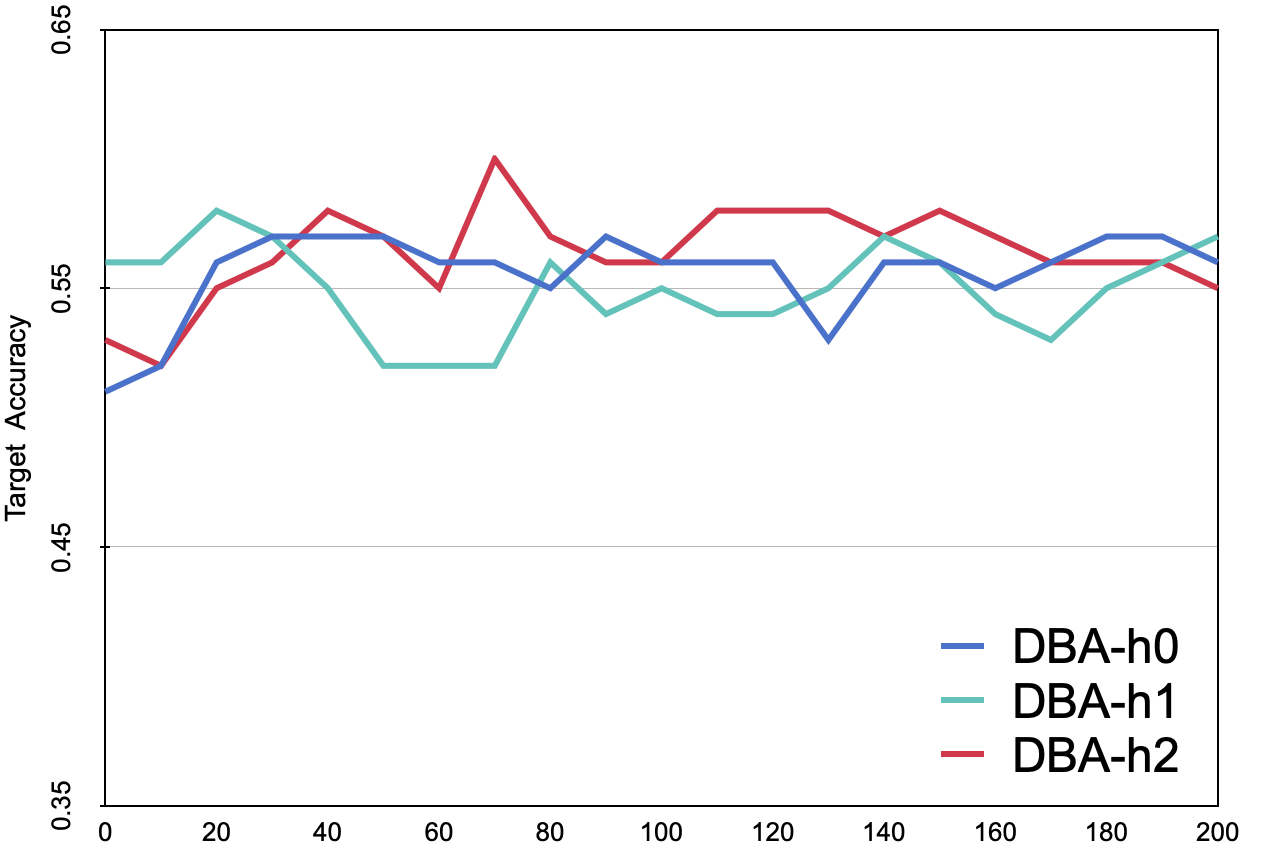}}
  \subfloat[\diversitycontrib+ \penaltycontrib]{\includegraphics[scale=0.2]{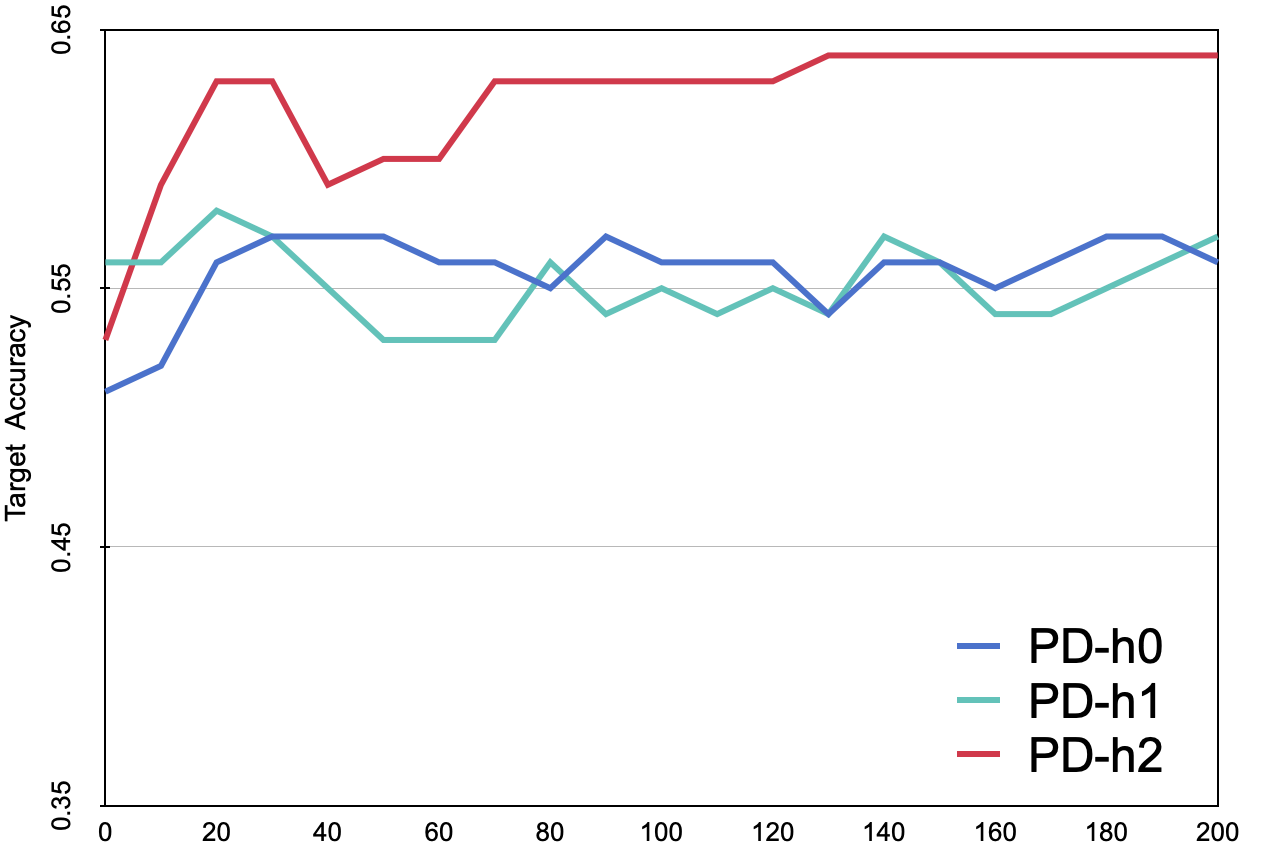}}
  \caption{\small{{\bf Diversity of the target hypotheses during adaptation} based on the choice of feature extraction backbone on S $\rightarrow$ T (LIDC). (a): three hypotheses that share feature extractor (\sharedb). (b): three hypotheses with separate feature extractors (\separateb). (c): \diversitycontrib with three independent hypotheses. (d): \model= \diversitycontrib+ \penaltycontrib with three independent hypotheses (\diversitycontrib).}} \label{fig:learning_curve}
\end{figure*}

To further examine the diversity of different ways of constructing the ensemble in the target adaption phase, we show one learning curve example for each model using fixed anchor selection (HDMI objective) in Figure~\ref{fig:learning_curve}. As expected, \sharedb leads to the lowest diversity. Introducing separate backbones (\separateb and \diversitycontrib) induces diversity that leads to an increase in performance. 
Furthermore, adding a regularizer as the combination of \diversitycontrib and \penaltycontrib seems to enable one of the hypotheses to find its way out of a local minimum, underscoring the synergistic impact of \modelname. 

\subsubsection{\penaltycontrib mitigates the negative influence of weak hypotheses}\label{seq:anchor_selection}
We studied the effect of anchor selection in the target hypothesis disparity regularization. Assuming source and target hypotheses have separate backbones, the performance of our model under fixed, random, ensemble (average), and \penaltycontrib anchor selection strategies are presented in Table~\ref{tab:anchor_comparison_lidc}. It can be observed from Table~\ref{tab:anchor_comparison_lidc} in several experiments, such as S $\rightarrow$ T, in the presence of weak performing hypothesis, while an ensemble anchor without WHP is subject to convergence towards weak hypotheses, random anchor selection might mitigate this issue partially. However, our results suggest that \penaltycontrib, through the penalization of outlier hypotheses, provides the most efficient protection against the negative impact of weak hypotheses by assigning them lower weights in the ensemble anchor.

Results on Office-31 dataset with three hypotheses indicate similar findings (see Table~\ref{tab:anchor_comparison_office_31}). Given three hypotheses, in all of the fixed, random, and ensemble anchor selection strategies, the impact of a weak hypothesis is inevitable in the overall performance. The results also indicate the instability of random strategy. For instance, in A $\rightarrow$ W, randomly choosing an anchor improved the performance in comparison with fixed selection, while in D $\rightarrow$ A, the model seems to converge toward the weak hypothesis. Our results on both natural and medical domains state that using \penaltycontrib helps to mitigate the effects of weak hypotheses.

\begin{table*}[!htb]
\centering
\caption{\small{Ablation study on different components of \model = \diversitycontrib + \penaltycontrib under covariate shift. Target accuracy ($\%$) on LIDC dataset using three hypotheses (source $\rightarrow$ target). The results are averaged over five different runs.}}\label{tab:diversity}
\begin{adjustbox}{width=1\textwidth}
\begin{tabular}{lccccccccccccc}
\toprule
Method & G$\rightarrow$P & G$\rightarrow$S & G$\rightarrow$T & P$\rightarrow$G & P$\rightarrow$S & P$\rightarrow$T & S$\rightarrow$G & S$\rightarrow$P & S$\rightarrow$T & T$\rightarrow$G & T$\rightarrow$P & T$\rightarrow$S &  Avg. \\
\midrule

\sharedb ~~+ Fixed & 67.1 & 66.6 & \textbf{64.6} & 65.4 & 64.2 & 54.6 & \textbf{66.2} & 65.1 & 54.6 & 60.7 & \textbf{60.0} & 58.7 & 62.3 \\ 

\separateb ~~+ Fixed & 66.7 & 64.4 & 57.3 & 64.5 & 63.3 & 53.2 & 65.2 & 66.4 & 55.5 & 60.4 & 59.9 & 57.8 & 61.2 \\

\diversitycontrib+ Fixed & 68.1 & \textbf{66.5} & 60.7 & 64.5 & 64.2 & \textbf{55.5} & 65.4 & \textbf{67.8} & 58.6 & 59.7 & 58.5 & 59.1 & 62.4 \\


\midrule
\diversitycontrib+\penaltycontrib & \textbf{68.9} & 65.9 & 60.9 & \textbf{65.6} & \textbf{65.6} & 54.8 & 65.8 & 66.9 & \textbf{66.6} & \textbf{61.2} & 59.6 & 59.6 & \textbf{63.5} \\ 

\bottomrule
\end{tabular}
\end{adjustbox}
\end{table*}

\subsubsection{Penalized Diversity relies on the synergy of DBA and WHP}
We ablate different components of the proposed \modelname (\model) for test-time adaptation performance on the LIDC dataset in Table~\ref{tab:diversity}. Table~\ref{tab:diversity} shows that using a \textit{Fixed} anchor selection, as done in HDMI, can lead to catastrophic failure cases due to error accumulation towards a weak hypothesis, which deems \textit{Fixed} a poor choice for anchor selection.
It is important to note that the increase in diversity seen in \separateb and \diversitycontrib (Fig.~\ref{fig:diversity}) results in worse performance when proper regularization is lacking (\separateb + Fixed and \diversitycontrib + Fixed). It is only when \penaltycontrib is introduced that we can mitigate the probability of convergence towards weak hypotheses.

\subsubsection{Weighted MI mitigates label distribution shift}
From Table~\ref{tab:digit_label_shift}, we observe that the performance of SHOT and HDMI dropped by more than $20\%$ in $L_m$ experiment in comparison with the covariate shift only experiment ($N_l$) indicating the incapability of these models to perform under label distribution shift. Similarly, OSTAR and MARS which both designed to tackle label shift and unlike \model have access to the source data during adaptation, had more than $14\%$ drop in their performance in $L_m$ experiment.
While our target estimation obtained from pseudo-labels is prone to errors, it significantly mitigates the catastrophic impact of label distribution shift by only $8\%$ performance drop. The effect of our modified MI maximization is remarkable in $L_t$ experiment where there is only $0.2\%$ drop in the performance of \model in comparison with no label shift ($N_l$).
It should be noted that our earlier experiments showed that applying $W$ to both label entropy, $H(\hat{Y}^T)$, and conditional entropy, $H(\hat{Y}^T \mid X^T)$, of MI maximization is no better than applying it solely to the label entropy.

\subsubsection{Estimated class proportions via pseudo-labels closely represent true class proportions in UDA}
To evaluate the effect of weighted MI maximization in the performance of \model under label distribution shift, we compare the performance of \model with and without weighted MI maximization (\model-NWMI) on \textit{Minority-Class shift} experiment. In this experiment, we choose 5 classes out of 10 in the target domain and changed their proportions by $p=0.1$. Table~\ref{tab:pseudo_label_performance} demonstrates a significant improvement ($14\%$) on using estimated target class proportions under label distribution shift.

\begin{table}[t] 
\centering
\caption{\small{Target accuracy ($\%$) on digit datasets with minority-class ($L_m$) label distribution shift with $p=0.1$ (source $\rightarrow$ target). In these experiments, \model-NWMI refers to \model without the weighted MI maimization, and \model-T refers to \model with the true target class proportions as compared to class proportion generated via pseudo-labels.}}\label{tab:pseudo_label_performance}
\begin{adjustbox}{width=0.5\textwidth}
\begin{tabular}{lccccccc}
\toprule
Method  & M$\rightarrow$U & M$\rightarrow$N & U$\rightarrow$M & U$\rightarrow$N & N$\rightarrow$M & N$\rightarrow$U &  Avg. \\
\midrule
\model-NWMI & 65.9 & 46.5 & 61.3 & 25.3 & 68.5 & 62.7 & 55.0 \\
\model & 87.5 & 47.7 & 84.3 & 31.8 & 85.0 & 78.6 & 69.2 \\
\model-T & 92.4 & 52.4 & 85.1 & 31.8 & 87.1 & 80.7 & 71.6 \\
\bottomrule
\end{tabular}
\end{adjustbox}
\end{table}
We further experiment on \textit{Minority-Class shift} with the actual target class proportions. It can be observed from Table~\ref{tab:pseudo_label_performance} that the performance of \model using the estimated target class proportions (PD) is close to true target class proportions (PD-T) implying the effectiveness of pseudo-labeling.

\subsection{Calibration Analysis}
It has been shown that diverse ensemble models lead to the best-calibrated uncertainty estimators~\cite{lakshminarayanan2017simple}. To evaluate the effect of diversity in \model from the calibration perspective, we compute the uncertainty and calibration of \model with Brier score~\cite{brier1950verification} and Expected Calibration Error (ECE)~\cite{naeini2015ece} and compare it with two other unsupervised SFDA models.

For this experiment, we consider natural and synthetic datasets under covariate and label distribution shifts. It can be observed from Table~\ref{tab:calibration}, that \model performs well in terms of calibration metrics under covariate shift in natural dataset. We also compare the performance of SFDA models on digit datasets with covariate shift only and with both covariate and label distribution shifts. For the experiment with both covariate and label shifts, we compute the calibration metrics in \textit{Minority-Class Shift} with $p=0.1$. As seen from Table~\ref{tab:calibration}, changing the proportion of classes in the target domain not only negatively impacts the transferability of the other two unsupervised SFDA models but also worsens these models' calibration. However, \model with a weighted MI maximization performs significantly better in terms of both performance and calibration after the introduction of label shift.
\begin{table*}[!htb]
    \centering
    \caption{\small{Calibration estimations for source-free domain adaptation models on the target domains A $\rightarrow$ D, Office-31, and M $\rightarrow$ U from digit dataset. Here $N_l$ and $L_m$ represent covariate shift only and covariate shift plus label distribution shift respectively. For the label distribution shift, we consider \textit{Minority-Class shift} with $p=0.1$. ${\ast}$ indicates calibration results reported from the original paper.}}\label{tab:calibration}
    \begin{adjustbox}{width=0.9\textwidth}
    \begin{tabular}{lccccc}
    \toprule
       Model & Dataset & Shift & Target acc. & Brier Score $\downarrow$ & ECE $\downarrow$ \\
        \midrule
        SHOT~\cite{liang2020shot} & \multirow{3}{*}{A $\rightarrow$ D} & & 93.1 & 0.1246 & 0.0039 \\
        HDMI~\cite{lao2021hdmi}$^{\ast}$ & & $N_l$ & 94.4 & 0.0961 & 0.0031 \\
        \model & &  & 95.6 & \textbf{0.0771} & \textbf{0.0024} \\
        \midrule
        SHOT~\cite{liang2020shot} & \multirow{3}{*}{M $\rightarrow$ U}& & 88.9 & 0.2170 & 0.0072 \\
        HDMI~\cite{lao2021hdmi} & & $N_l$ & 95.2 & 0.0926 & 0.0030 \\
        \model & &  & 96.9 & \textbf{0.0567} & \textbf{0.0011} \\
        \midrule
        SHOT~\cite{liang2020shot} & \multirow{3}{*}{M $\rightarrow$ U} & & 57.1 & 0.8432 & 0.0279 \\
        HDMI~\cite{lao2021hdmi} & &  $L_m$ & 62.4 & 0.7417 & 0.0246 \\
        \model & &  & 87.5 & \textbf{0.2467} & \textbf{0.0082} \\
        \bottomrule
    \end{tabular}
    \end{adjustbox}
\end{table*}

\subsection{Sensitivity Analysis}
\subsubsection{\penaltycontrib is robust to hyper-parameter selection}
\begin{figure*}[!htb] 
  \centering
  \subfloat[Model sens. (LIDC)]{\includegraphics[scale=0.182]{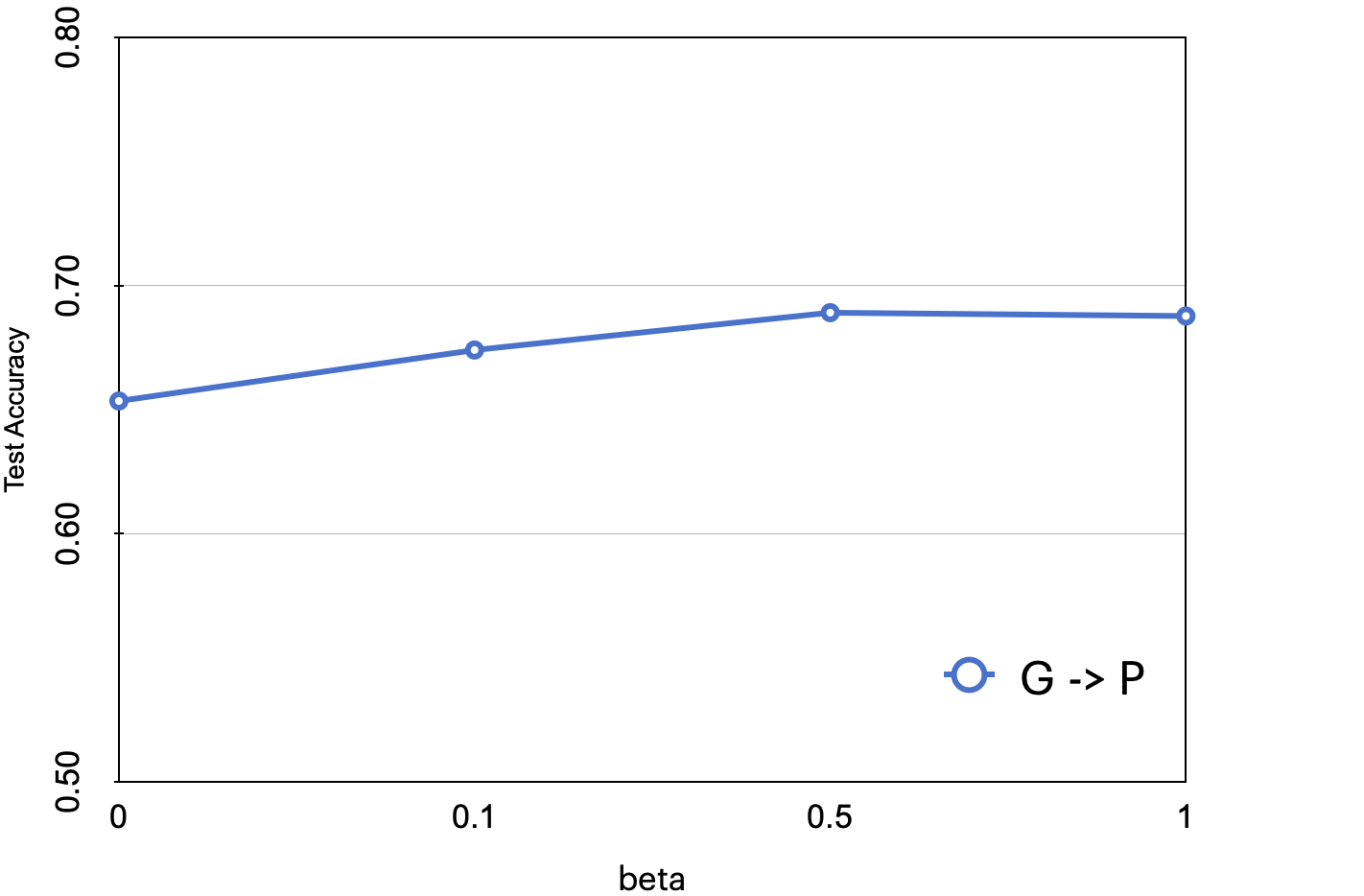}}
  \subfloat[Model sens. (Office-31)]{\includegraphics[scale=0.182]{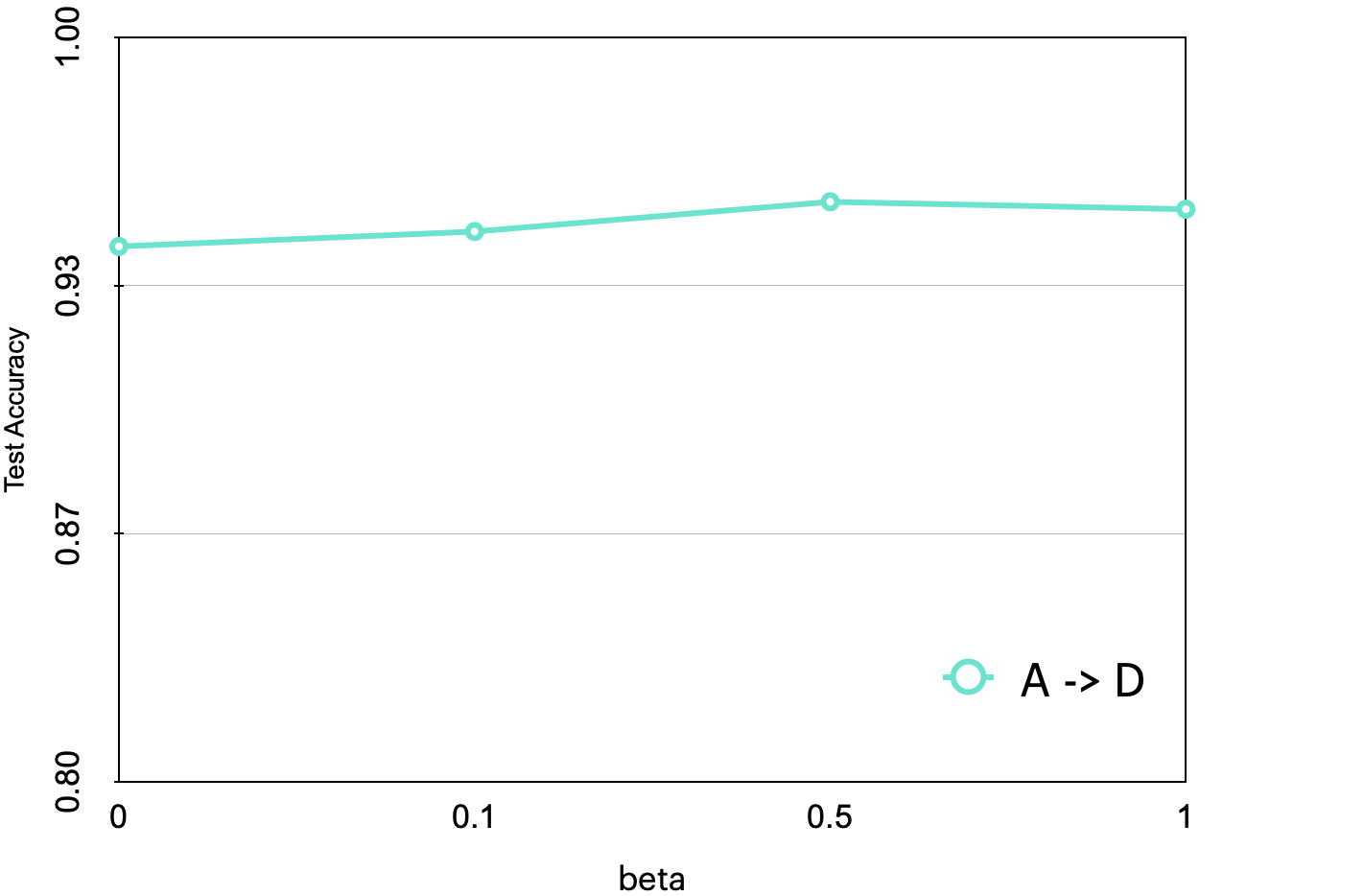}}
  \subfloat[USPS class dist.]{\includegraphics[scale=0.09]{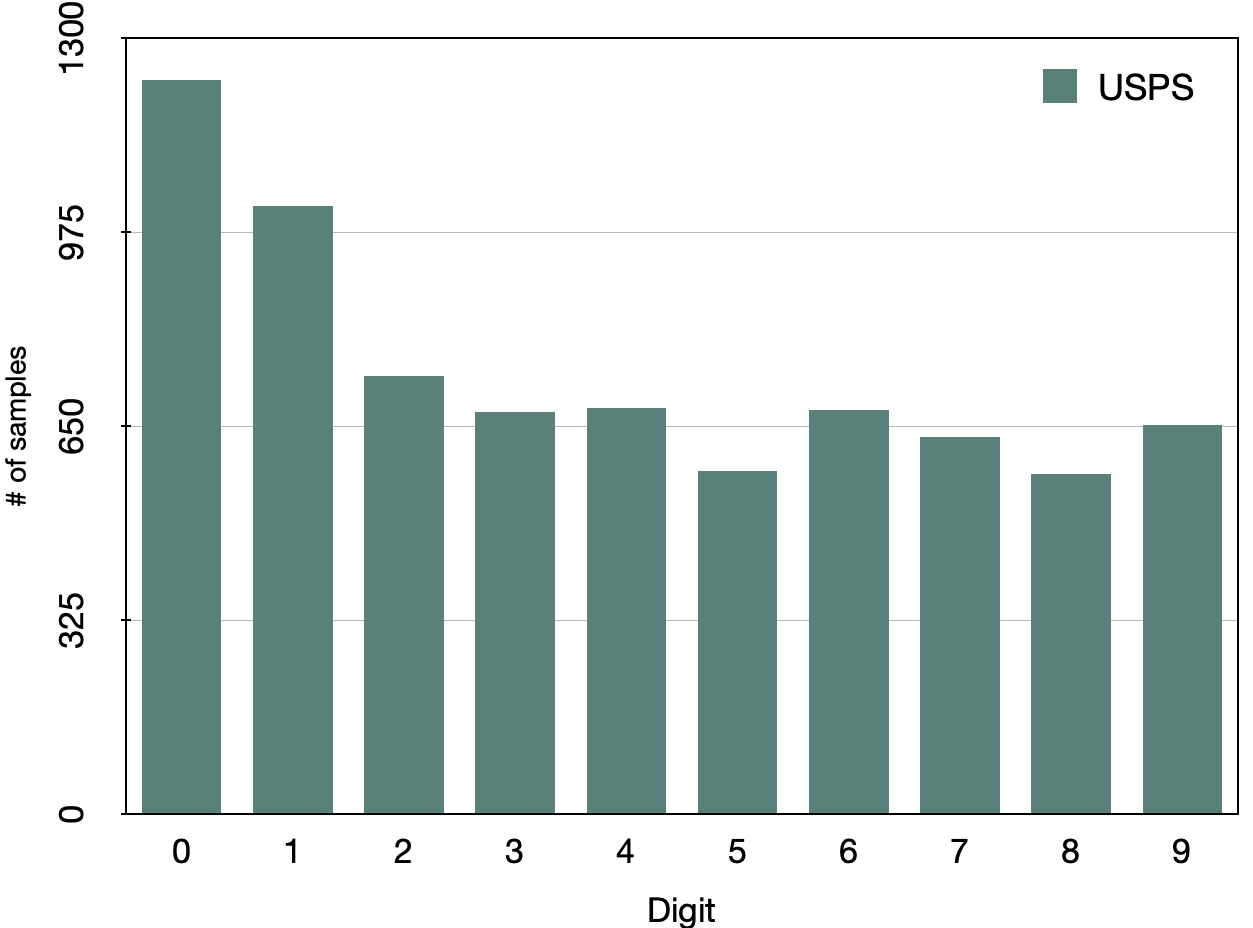}}
  \subfloat[MNIST class dist.]{\includegraphics[scale=0.09]{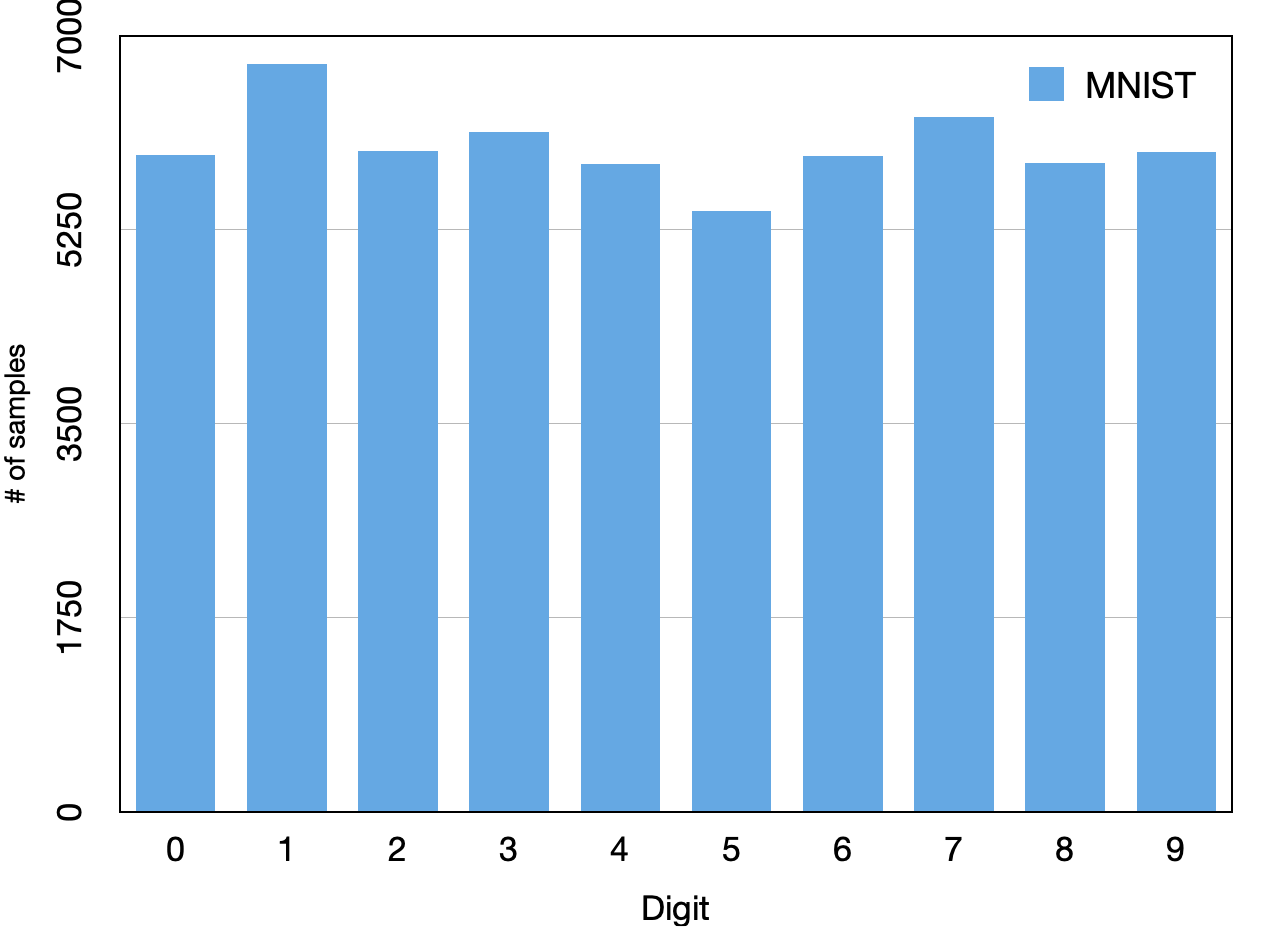}}
  \caption{\small{The two left plots show $\beta$ sensitivity in G$\rightarrow$P (LIDC) and A$\rightarrow$D (Office-31) with three hypotheses. The two right plots indicate class distribution in USPS and MNIST datasets. Note that the class distribution of MNIST-M is the same as MNIST.}}\label{fig:sensitivity_data_dist}
\end{figure*}

To investigate the sensitivity of our model to the hyperparameter $\beta$, we conduct a set of experiments on A$\rightarrow$D (Office-31) and G$\rightarrow$P (LIDC) with three hypotheses and summarize the results in Fig.~\ref{fig:sensitivity_data_dist}. For this experiment, we fix $\alpha=0.3$ for LIDC and $\alpha=1$ for Office-31. Setting $\beta=0$ reduces to solely maximizing mutual information. Fig~\ref{fig:sensitivity_data_dist} (a) and (b) show that introducing target training with \penaltycontrib improved the performance in comparison with mutual information maximization ($\beta=0$). It is seen from the figure that despite the difference between the two domains (natural and medical), increasing the \penaltycontrib contribution in target training improves the performance.

\subsection{Ablation Study}
\subsubsection{Choice of Different Architectures on PD}\label{sec:ablation_study}
\begin{figure*}[t] 
  \centering
  \subfloat[\diversitycontrib (A1)]{\includegraphics[scale=0.2]{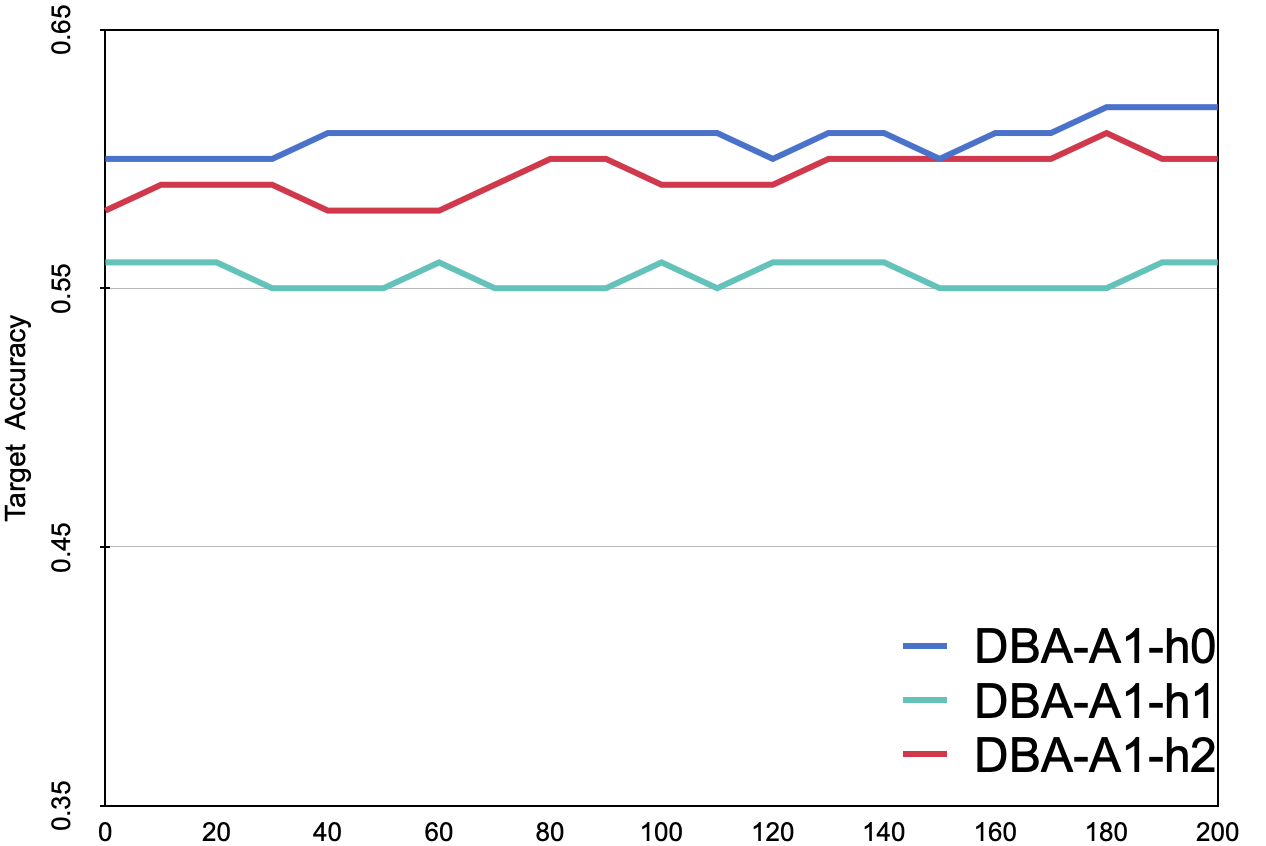}}
  \subfloat[\model (A1)]{\includegraphics[scale=0.2]{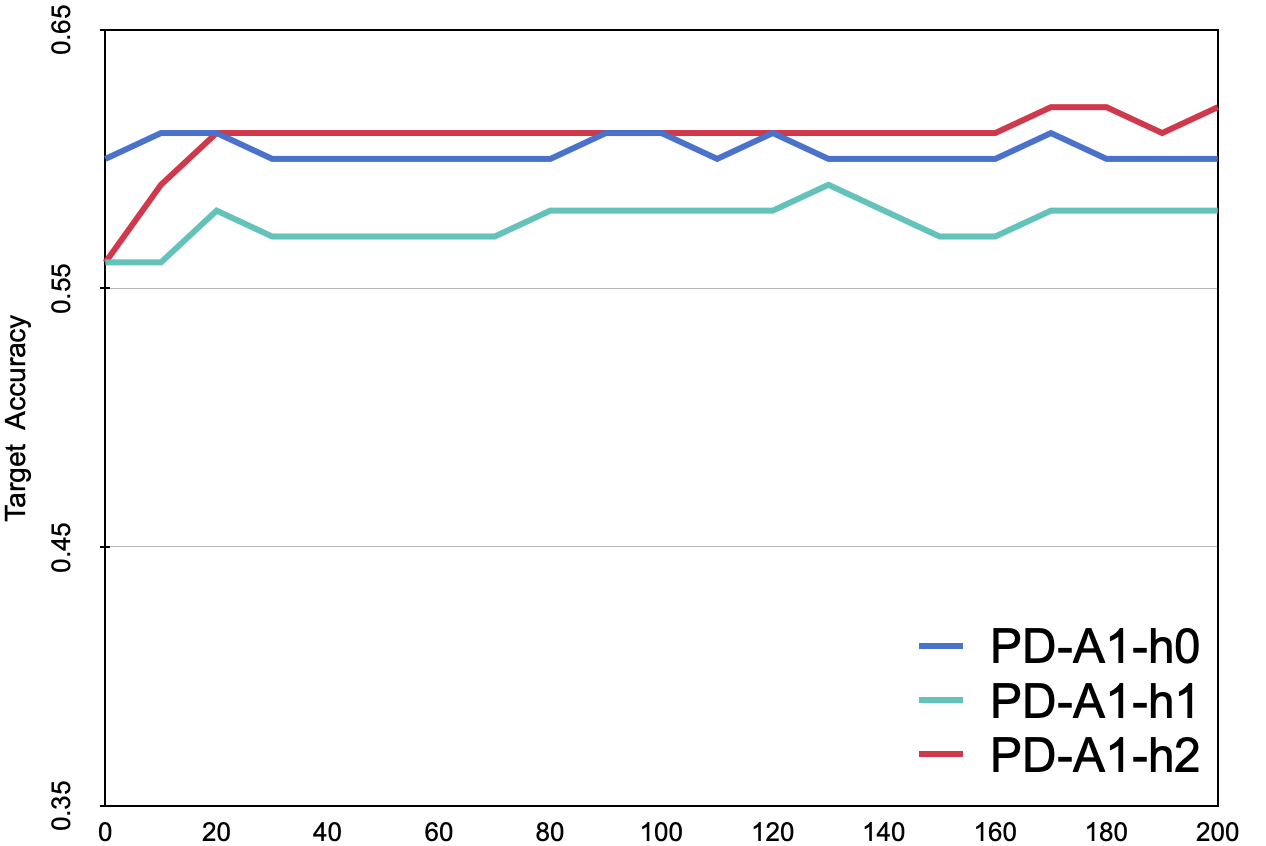}}
  \subfloat[\diversitycontrib (A2)]{\includegraphics[scale=0.2]{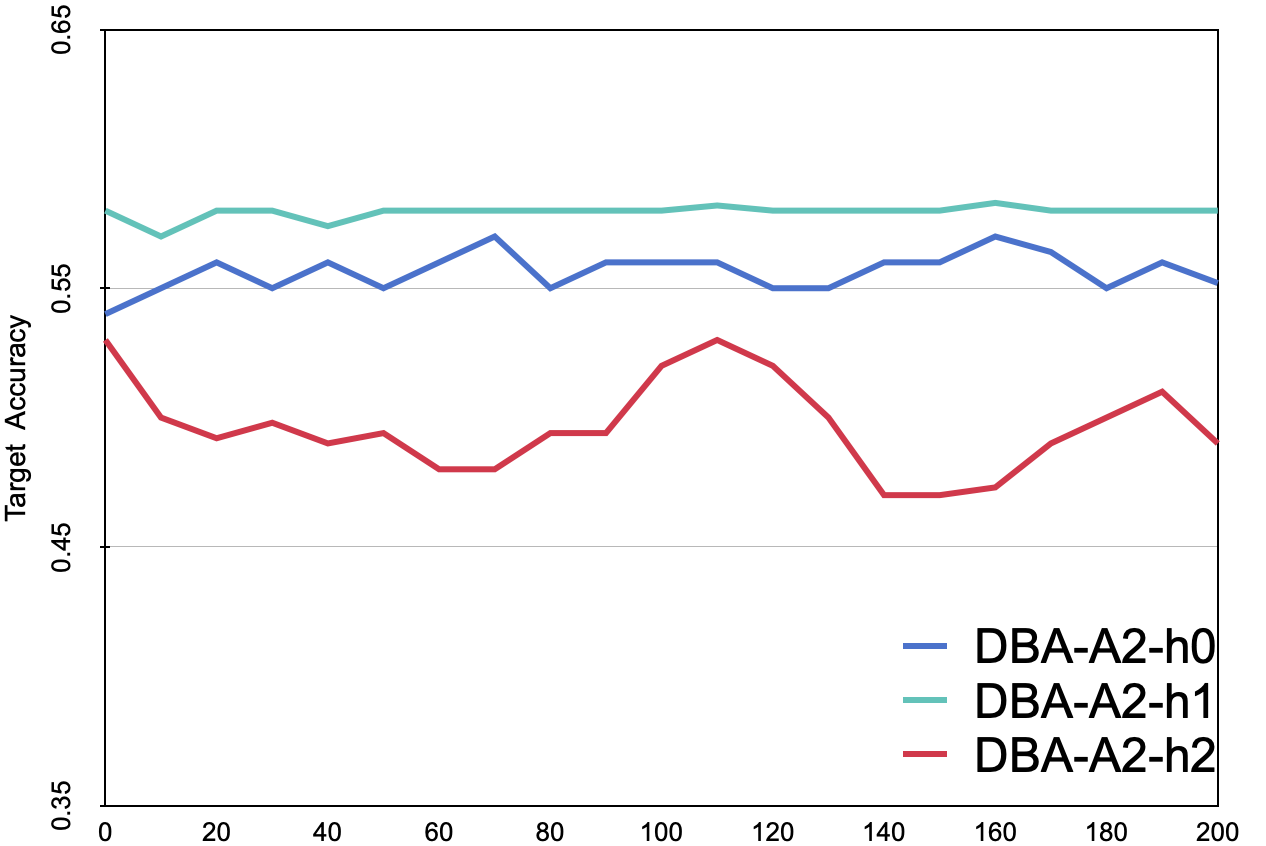}}
  \subfloat[\model (A2)]{\includegraphics[scale=0.2]{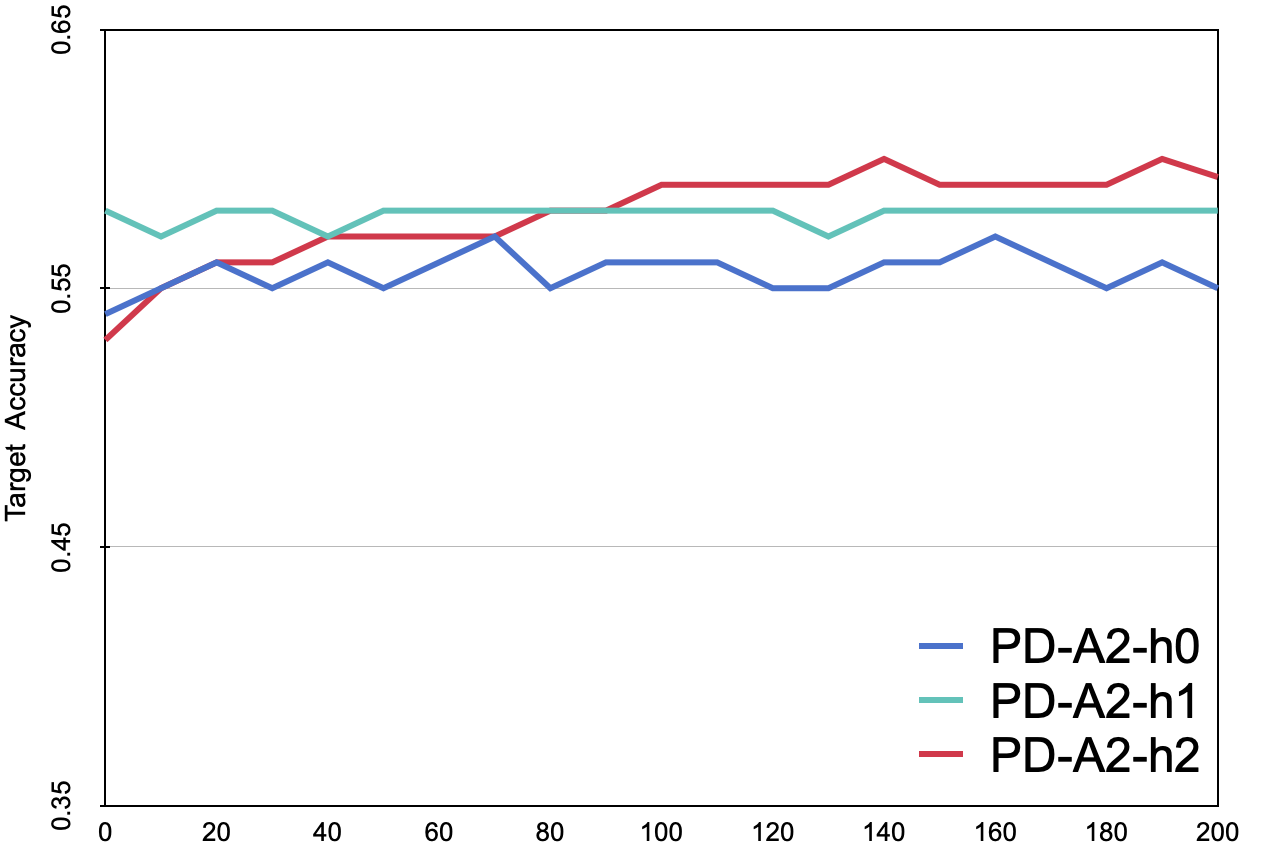}}
  \caption{\small{{\bf Diversity of the target hypotheses during adaptation using various architecture choices} on T $\rightarrow$ G from LIDC dataset. A1 represents the first choice of different backbone architectures including different depths of ResNet \{10, 18, 10\}. A2 represents the second choice of different backbone architectures including ResNet10, ResNet18, and SqueezeNet1.0.  (a): \diversitycontrib with three hypotheses and fixed anchor on the first architecture (A1). (b): \model with three hypotheses on the first architecture (A1). (c): \diversitycontrib with three hypotheses and fixed anchor on the second architecture (A2). (d): \model with three hypotheses on the second architecture (A2).}} \label{fig:learning_curve_dif_arch}
\end{figure*}

We study the effects of different architectural designs on the performance of \model as well as the diversity of the hypotheses. We compare two different choices of architectures for \diversitycontrib. In this study, we simply consider different depths of a network as different backbones of \model (refers as A1). However, to investigate the performance of \model under totally different architectures, we consider a combination of SqueezeNet~\cite{iandola2016squeezenet} and ResNet in \model (refers as A2) with three hypotheses on LIDC. From Figure~\ref{fig:learning_curve_dif_arch}, we can observe that three hypotheses with entirely different architectures also improve the diversity. However, \diversitycontrib without a proper regularizer creates uncontrolled diversity as shown in Figure~\ref{fig:learning_curve_dif_arch}(c).
The experimental results presented in Table~\ref{tab:dif_arch} show that imposing diversity on the model through entirely different architectural designs (i.e. A2) also leads to improvement in comparison with \sharedb with similar backbone architectures (from $62.3\%$ to $62.8\%$).
\begin{table*}[!htb]
\centering
\caption{\small{Ablation study on different architectural choices. Target accuracy ($\%$) on LIDC dataset (source $\rightarrow$ target) under covariate shift with two different choices of architectures; A1 and A2. The results are averaged over five different runs.}}\label{tab:dif_arch}
\begin{adjustbox}{width=1\textwidth}
\begin{tabular}{lccccccccccccc}
\toprule
Method & G$\rightarrow$P & G$\rightarrow$S & G$\rightarrow$T & P$\rightarrow$G & P$\rightarrow$S & P$\rightarrow$T & S$\rightarrow$G & S$\rightarrow$P & S$\rightarrow$T & T$\rightarrow$G & T$\rightarrow$P & T$\rightarrow$S &  Avg. \\
\midrule

\model (A1) & \textbf{68.9} & \textbf{65.9} & 60.9 & \textbf{65.6} & \textbf{65.6} & 54.8 & \textbf{65.8} & \textbf{66.9} & \textbf{66.6} & \textbf{61.2} & \textbf{59.6} & \textbf{59.6} & \textbf{63.5} \\ 

\model (A2) & 66.2 & 65.3 & \textbf{64.3} & 63.6 & 65.2 & \textbf{59.1} & 64.7 & \textbf{66.9} & 61.1 & 60.6 & 58.4 & 58.2 & 62.8 \\

\bottomrule
\end{tabular}
\end{adjustbox}
\end{table*}

\section{Related Work}
Transfer learning approaches can be divided into data-driven and model-driven approaches. Data-driven approaches such as instance weighting~\cite{zadrozny2004learning,bickel2007discriminative} and domain adaptation models such as DAN~\cite{long2015learning}, DANN~\cite{ganin2015unsupervised} and MDD~\cite{zhang2019bridging} assume to have direct access to the source data during the knowledge transfer. To mitigate transfer learning models' privacy and storage concerns, source-free domain adaptation (SFDA) approaches (also known as model-driven or hypothesis transfer learning) are proposed. SFDA is a transfer learning strategy where a model trained on the source domain incorporates the learning procedure of the target domain. It was first introduced by \cite{kuzborskij2013stability} where the access to the source domain was only limited to a set of hypotheses induced from it, unlike domain adaptation, where both source and target domains are used to adapt the source hypothesis to the target domain.

\textbf{Source-free Domain Adaptation} has been investigated from both practical and theoretical points of view in computer vision applications. Several studies have analyzed the effectiveness of SFDA on various specific ML algorithms~\citep{kuzborskij2013stability,kuzborskij2017fast,wang2015generalization}, while others proposed more generally applicable frameworks~\cite{fernandes2019hypothesis,du2017hypothesis}. These studies can be divided based on the availability of labeled data in the target domain. Most previous studies considered the supervised SFDA setting (labeled target domain)~\cite{kuzborskij2013stability,kuzborskij2017fast,wang2015generalization,fernandes2019hypothesis,du2017hypothesis}, while unsupervised SFDA (uSFDA) (unlabeled target domain) has only recently gained interest~\cite{liang2020shot,lao2021hdmi}. SFDA models mostly rely on a single hypothesis to transfer knowledge to the target domain. \citet{lao2021hdmi} showed that using a single hypothesis for uSFDA is prone to overfitting the target domain and causes catastrophic forgetting of the source domain. They were the first to propose using multiple hypotheses to mitigate this effect. More recently~\citet{wang2022dipe} proposed a novel way to tackle the SFDA problem by finding domain-invariant parameters rather than domain-invariant features in the model. 

\textbf{Ensemble Models}
Recently, deep neural network calibration gained considerable attention in the machine learning research community. Previous studies explored the effect of Monte Carlo dropout~\cite{kingma2015variational,gal2016dropout} and variational inference methods~\cite{maddox2019simple}. However, it has been shown that the best-calibrated uncertainty estimators can be achieved by neural network ensembles~\cite{lakshminarayanan2017simple,ovadia2019can,Ashukha2020Pitfalls}. The importance of well-calibrated models becomes more important under the presence of dataset shift.
The success of ensemble models is mainly related to the diversities present between their members. Ensemble diversity has been widely investigated in the literature \cite{brown2004diversity,liu1999ensemble}. Improving diversity in neural network ensembles has become a focus in recent work. \citet{stickland2020diverse} suggest augmenting each member of an ensemble with a different set of augmented input to increase the diversity among members. While a few recent studies propose deep ensemble models based on different neural network architectures to ensure diversity~\cite{antoran2020depth,zaidi2020neural}. 

Recently~\citet{pagliardini2022agree} suggest that encouraging diversity between the ensemble predictions helps to generalize in the OOD setting by increasing disagreements and uncertainties over out-of-distribution samples. Whereas~\citet{lee2022diversify} introduced an ensemble of multiple hypotheses with shared feature extractors and separate classifier heads to generalize in the presence of spurious features. They proposed to increase diversity among the classifiers through mutual information minimization over the hypotheses predictions on unlabeled target data. Our work is different from~\cite{pagliardini2022agree,lee2022diversify} in a sense that (i) to increase diversity, \model does not need a carefully selected set of target samples unlike both~\cite{pagliardini2022agree,lee2022diversify}, (ii) different from~\cite{pagliardini2022agree} that limits the model to have a smaller or equal number of hypotheses than the total number of classes, we have freedom over the number of hypotheses in our model, and (iii) \penaltycontrib mitigate weak hypotheses to improve overall performance without requiring access to labeled target samples as opposed to the active query strategy presented in~\cite{lee2022diversify}. Despite its performance, \model also has its own limitation. Its diversification and penalization approaches force \model to be more effective with an ensemble of at least three hypotheses. 

\textbf{Label Distribution Shift} Many domain adaptation studies focus only on covariate shift. Despite the negative impact of label distribution shift in transferring knowledge, it has been mostly neglected. Learning domain-invariant representations and using estimated class ratios between domains as importance weights in the training loss became a dominant strategy for many recent practices~\cite{gong2016domain,tachet2020domain,shui2021aggregating}. \citet{rakotomamonjy2022mars} proposed MARS to learn domain-invariant representations with sample re-weighting. Several studies attempt to benefit optimal transport (OT) to find a transport function from source to target with a minimum cost. \citet{kirchmeyer2022ostar} proposed a reweighing model which maps pretrained representations using OT. The key difference between these models and our modified MI solution is that they all assume accessing the source data during adaptation and their reweighing strategies are mainly based on source and target ratios.

\section{Conclusion}
This paper shows the benefits of increasing diversity in unsupervised source-free domain adaptation. We increased diversity by introducing separate feature extractors with Distinct Backbone Architectures (\diversitycontrib) across hypotheses. With the support of experiments on various domains, we show that diversification must be accompanied by proper Weak Hypothesis mitigation through Penalization (\penaltycontrib). Our proposed Penalized Diversity (\model) stems from the synergy of \diversitycontrib and \penaltycontrib. We further modified MI maximization in the objective of \model to account for the label shift problem. Our experiments on natural, synthetic, and medical benchmarks demonstrate how it improves upon the relevant baselines under different distributional shifts. As for future work, we would like to investigate other ways to promote diversity in the feature space of SFDA models.



\bibliography{LaTeX/aaai23}

\clearpage

\appendix
\section{Appendix}\label{sec:A1}

\begin{table*}[t]
\centering
\caption{\small{The confidence interval (CI) on each domain using the Source-only model on LIDC dataset with covariate shift.}}\label{tab:ci}
\begin{adjustbox}{width=0.9\textwidth}
\begin{tabular}{lcccccccccccc}
\toprule
Source & ~~ & Run & ~~ & G & ~~ & P & ~~ & S & ~~ & T \\
\midrule
  & ~~ & S1 & ~~ & -- & ~~ & \textbf{(-0.09, 2.95)} & ~~ & (-5.66, -0.20) & ~~ & (17.63, 21.69) \\
  & ~~ & S2 & ~~ & -- & ~~ & (-4.96, -1.98) & ~~ & \textbf{(-1.71, 3.85)} & ~~ & (12.53, 16.59) \\
G & ~~ & S3 & ~~ & -- & ~~ & (0.67, 3.73) & ~~ & (-5.53, -0.05) & ~~ & (18.18, 22.24) \\
  & ~~ & S4 & ~~ & -- & ~~ & (0.36, 3.40) & ~~ & (-6.15, -0.73) & ~~ & (18.10, 22.15) \\
  & ~~ & S5 & ~~ & -- & ~~ & (3.85, 6.95) & ~~ & (1.22, 6.90) & ~~ & (19.10, 23.14) \\
\midrule
  & ~~ & S1 & ~~ & \textbf{(-2.10, 0.14)} & ~~ & -- & ~~ & \textbf{(-3.32, 0.74)} & ~~ & (6.10, 12.62) \\
  & ~~ & S2 & ~~ & (3.15, 6.15) & ~~ & -- & ~~ & (6.50, 10.48) & ~~ & (15.25, 21.73) \\
P & ~~ & S3 & ~~ & (0.76, 3.82) & ~~ & -- & ~~ & (1.89, 5.89) & ~~ & (11.13, 17.63) \\
  & ~~ & S4 & ~~ & (-3.53, -0.51) & ~~ & -- & ~~ & \textbf{(-1.26, 2.70)} & ~~ & (12.27, 18.77) \\
  & ~~ & S5 & ~~ & (3.33, 6.45) & ~~ & -- & ~~ & (1.53, 5.57) & ~~ & (10.21, 16.72) \\
\midrule
  & ~~ & S1 & ~~ & (-3.17, -3.13) & ~~ & \textbf{(-0.00, 5.70)} & ~~ & -- & ~~ & (2.39, 8.85) \\
  & ~~ & S2 & ~~ & (1.07, 4.11) & ~~ & (-7.61, -2.27) & ~~ & -- & ~~ & (5.57, 12.01) \\
S & ~~ & S3 & ~~ & (1.25, 4.27) & ~~ & (-6.39, -1.05) & ~~ & -- & ~~ & (9.05, 15.51) \\
  & ~~ & S4 & ~~ & (-6.10, -3.05) & ~~ & (-10.85, -5.45) & ~~ & -- & ~~ & (5.05, 11.55) \\
  & ~~ & S5 & ~~ & (-5.42, -2.36) & ~~ & (-5.72, -0.12) & ~~ & -- & ~~ & \textbf{(-5.33, 0.97)} \\
\midrule
  & ~~ & S1 & ~~ & (18.93, 21.97) & ~~ & (17.57, 23.33) & ~~ & (18.44, 22.46) & ~~ & -- \\
  & ~~ & S2 & ~~ & (18.93, 21.97) & ~~ & (71.57, 23.33) & ~~ & (18.44, 22.46) & ~~ & -- \\
T & ~~ & S3 & ~~ & (22.36, 25.36) & ~~ & (21.0, 26.72) & ~~ & (21.87, 25.85) & ~~ & -- \\
  & ~~ & S4 & ~~ & (17.79, 20.85) & ~~ & (16.44, 22.20) & ~~ & (17.31, 21.33) & ~~ & -- \\
  & ~~ & S5 & ~~ & (17.79, 20.85) & ~~ & (16.44, 22.20) & ~~ & (17.31, 21.33) & ~~ & -- \\
\bottomrule
\end{tabular}
\end{adjustbox}
\end{table*}

\subsection{Experimental Setup}
For the medical dataset, we used different depths of 3D-ResNet as the hypothesis backbone, mainly ResNet10 and ResNet18. Each backbone $\{\phi_i\}_{i=1}^M$ is then followed by a set of fully connected layers, Batch-Norm, ReLU activation function, and Dropout referred to as the bottleneck layer. We used 512 as the dimension of extracted features. For the classifier $\{f_i\}_{i=1}^M$, we used a shallow neural network with two fully connected layers, followed by a ReLU activation function, and Dropout. Each model trained for $200$ iterations with batch size $32$ and learning rate $1e-4$ and AdamW optimizer on the source domain. We used the same configuration for the target training except the learning rate was decreased to $1e-5$. For the experiments with 3 hypotheses, we used ResNet $\{10, 18, 10\}$, indicating the depth of each feature extractor. 
For the baselines, we used the same configuration with ResNet18 as their backbone whether they have single or multiple hypotheses. $\alpha = 0.3$ and $\beta = 0.5$ are used for the target training.

For natural images datasets, we use different depths of ResNet~\cite{he2016deep} pre-trained on ImageNet~\cite{russakovsky2015imagenet} as the backbone of our feature extractors. Specifically, ResNet of depths $\{34, 50, 34\}$ for Office-31 and Office-Home and ResNet of depths $\{50, 101, 110\}$ for VisDA-C are chosen as the depths of $\{g_i\}_{i=1}^{M}$ for $M=3$ hypotheses. The same bottleneck layer as our experiments on medical data is used for the experiments on natural images. We followed similar hyperparameters as~\citep{liang2020shot} for synthetic digit datasets with different depths on \model. 
For both natural and synthetic datasets, we trained the source hypotheses for 5k iterations, with learning rate $3e-4$ and batch size of $64$. Target hypotheses are trained for 20k iterations with learning rate $3e-4$ and batch size of 64. We used SGD optimizer for both source and target training. We used $\alpha = 1$ and $\beta = 0.5$ for the target training objective function.

For the experiments on digit datasets, we used $p=0.1$ as the probability value of changing the chosen class, i.e. a class with $1000$ samples in the target domain will be reduced to $100$ samples in the new shifted domain.

\subsection{Statistical Analysis on LIDC Covariate Shift}
The Lung Image Database Consortium (LIDC)~\cite{armato2011lidc} consists of diagnostic and lung cancer screening thoracic computed tomography (CT) scans. The images are captured in multiple institutions with imaging devices produced by four different manufacturers. It has been shown that images from different institutions as well as hardware differences in data acquisition devices are susceptible to domain shift (also known as covariate shift)~\cite{guan2021domain,stacke2019closer,karani2018lifelong}. We suggested dividing the LIDC dataset into four sub-datasets based on the imaging device manufacturer. Each of these sub-datasets introduces one domain. Aside from the results presented in Table~\ref{tab:results_lidc} of the paper (Sec.~\ref{sec:lidc_result}, results on LIDC under covariate shift) obtained by the Source-only model, this section provides a statistical argument for the existence of a covariate shift in our suggested approach.

Assuming that each of the sub-datasets (i.e. domain) is different from the others introducing a comparative observational study or experiment, we assess the difference between proportions or means of each two experiments.

Given the performance of the Source-only model on each domain within five different runs, we computed the confidence interval (CI) between each two population proportions using Eq.~\ref{eq:ci}.

\small{
\begin{equation}\label{eq:ci}
    CI = \text{P difference} \pm \text{SE for Difference} \\
\end{equation}}

\noindent and
\begin{equation}
\begin{split}
    \text{P difference} &= pr_{d_1} - pr_{d_2}\\
    \text{SE for Difference} &= \sqrt{(\text{SE}_{p_1})^2 + (\text{SE}_{p_2})^2}\\
\end{split}
\end{equation}

\noindent where $SE$ for proportion $p_i$ defines standard error and it is computed as follows:

\begin{equation*}
    SE_{p_i} = \sqrt{\frac{pr_{d_i}(1-pr_{d_i})}{N_{p_i}}}
\end{equation*}

\noindent where $N_{p_i}$ indicates the total number of samples in each proportion/domain, and $pr_{d_i}$ is the accuracy of Source-only model trained on the proportion $p_i$.

If two CIs do not overlap, then it can be said that there is a statistically significant difference between the two populations. In another word, if the confidence interval for the difference does not contain zero, we can confirm the existence of covariate shift between two domains.

The highlighted values in Table~\ref{tab:ci} indicate overlaps between two domains on a particular run. As seen from Table~\ref{tab:ci}, the highest domain shift is observed between T and the other domains. These findings are also aligned with the results reported in the paper on the LIDC dataset, where the lowest performance is obtained where T is either source or target domain (see Table~\ref{tab:results_lidc}). Since in nearly all the experiments, at least four out of five experiments show no overlaps, we conclude that our suggested approach to creating sub-datasets indeed maintains domain shift.


\end{document}